\documentclass[10pt,twocolumn,letterpaper]{article}

\usepackage{iccv}
\usepackage{times}
\usepackage{epsfig}
\usepackage{graphicx}
\usepackage{amsmath}
\usepackage{amssymb}
\usepackage{subfig}
\usepackage{tabularx}
\usepackage{booktabs}
\usepackage{multirow}
\usepackage{enumitem}
\usepackage{multicol}
\usepackage{csquotes}
\usepackage{caption}
\usepackage[dvipsnames, table]{xcolor}
\usepackage{color}
\usepackage{rotating}
\usepackage{floatrow}
\usepackage{placeins}
\usepackage{array, makecell}
\usepackage[flushleft]{threeparttable} 
\usepackage{csquotes}
\usepackage{cite}
\usepackage{algorithm}
\usepackage{algpseudocode}
\usepackage{amstext} 
\usepackage{array}   
\newcolumntype{M}{>{$}c<{$}}

\usepackage[pagebackref=true,breaklinks=true,colorlinks,bookmarks=false]{hyperref}

\renewcommand{\mkbegdispquote}[2]{\itshape}

\algblock{Input}{EndInput}
\algnotext{EndInput}
\algblock{Output}{EndOutput}
\algnotext{EndOutput}
\newcommand{\Desc}[2]{\State \makebox[2em][l]{#1}#2}

\newcommand{\ra}[1]{\renewcommand{\arraystretch}{#1}}
\newlength{\tempdima}
\newcommand{\rowname}[1]
{\rotatebox{90}{\makebox[\tempdima][c]{\textbf{#1}}}}

\newcommand{\bx}{\mathbf{x}}
\newcommand{\bz}{\mathbf{z}}

\newcommand{\IR}{\mathbb{R}}
\newcommand{\IE}{\mathbb{E}}
\newcommand{\EL}{\mathcal{L}}
\newcommand{\EF}{\mathcal{F}}

\newcommand{\ED}{\mathcal{D}}

\newcommand{\EP}{\mathcal{P}}
\newcommand{\EPur}{\mathcal{P}ur}

\newcommand{\EG}{\mathcal{G}}

\DeclareRobustCommand{\eg} {\textit{e}.\textit{g}.}
\DeclareRobustCommand{\ie}{\textit{i}.\textit{e}.}
\DeclareRobustCommand{\etal}{\textit{et~al.}~}

\newcommand{\Section}[1]{\vspace{-1mm} \section{#1} \vspace{1mm}}
\newcommand{\SubSection}[1]{\vspace{-1mm} \subsection{#1} \vspace{-1mm}}

\newcommand{\Paragraph}[1]{\vspace{1.25mm}\noindent\textbf{#1.}\hspace{0.5mm}}

\iccvfinalcopy 

\def\iccvPaperID{10029} 

\begin{document}

\title{FaceGuard: A Self-Supervised Defense Against Adversarial Face Images}

\author{Debayan Deb, Xiaoming Liu, Anil K. Jain\\
Department of Computer Science and Engineering,\\
Michigan State University, East Lansing, MI, 48824\\
{\tt\small{\{debdebay, liuxm, jain\}@cse.msu.edu}}}

\twocolumn[{%
\renewcommand\twocolumn[1][]{#1}%
\maketitle
\begin{center}
    \centering
    \captionsetup{font=footnotesize}
    \includegraphics[width=0.95\linewidth]{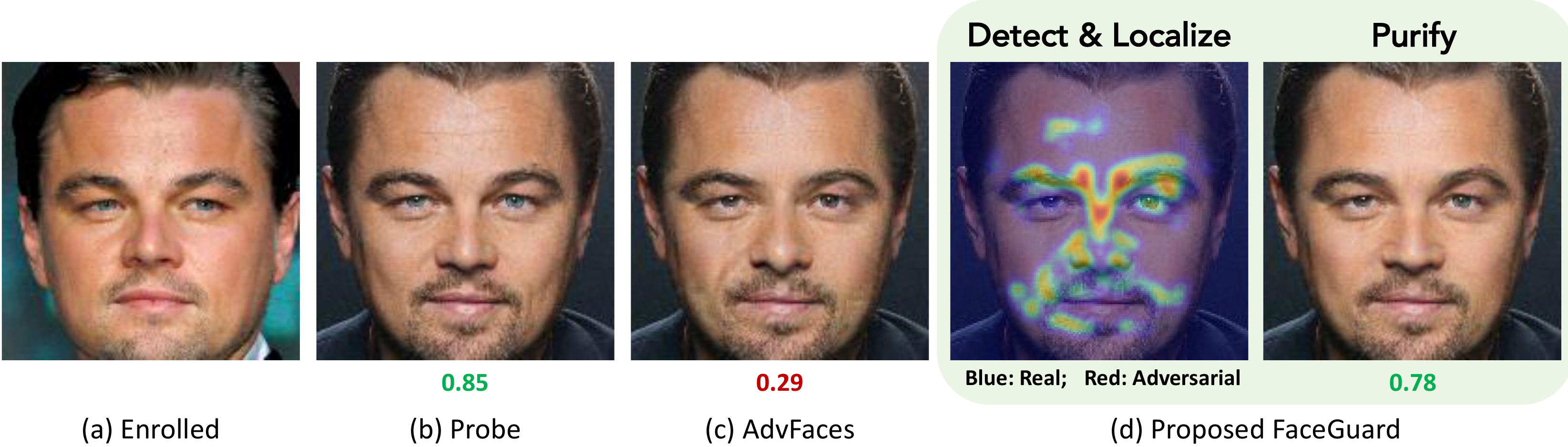}\vspace{-2mm}
    \captionof{figure}{Leonardo DiCaprio's real face photo (a) enrolled in the gallery and (b) his probe image\protect\footnotemark; (c) Adversarial probe synthesized by a state-of-the-art (SOTA) adversarial face generator, AdvFaces\cite{advfaces}; (d) Proposed adversarial defense framework, namely~\emph{FaceGuard} takes (c) as input, detects  adversarial images, localizes perturbed regions, and  outputs a ``purified" face devoid of adversarial perturbations. A SOTA face recognition system, ArcFace, fails to match Leonardo's adversarial face (c) to  (a), however, the purified face can successfully match to (a). Cosine similarity scores ($\in[-1,1]$) obtained via ArcFace~\cite{arcface} are shown below the images. A score above $\textbf{0.36}$ (threshold @ $0.1\%$ False Accept Rate) indicates that two faces are of the same subject.}
    \label{fig:frontpage}
\end{center}%
}]

\begin{abstract}
\vspace{-1em}Prevailing defense schemes against adversarial face images tend to overfit to the perturbations in the training set and fail to generalize to unseen adversarial attacks. We propose a new self-supervised adversarial defense framework, namely FaceGuard, that can automatically detect, localize, and purify a wide variety of adversarial faces without utilizing pre-computed adversarial training samples. During training, FaceGuard automatically synthesizes challenging and diverse adversarial attacks, enabling a classifier to learn to distinguish them from real faces. Concurrently, a purifier attempts to remove the adversarial perturbations in the image space. Experimental results on LFW, Celeb-A, and FFHQ datasets show that FaceGuard can achieve $99.81\%$, $98.73\%$, and $99.35\%$ detection accuracies, respectively, on six \textbf{unseen} adversarial attack types. In addition, the proposed method can enhance the face recognition performance of ArcFace from $34.27\%$ TAR @ $0.1\%$ FAR under no defense to $77.46\%$ TAR @ $0.1\%$ FAR.
 Code, pre-trained models and dataset will be publicly available.
\end{abstract}

\Section{Introduction}
With the advent of deep learning and availability of large  datasets, Automated Face Recognition (AFR) systems have achieved impressive recognition rates~\cite{nist_ongoing}. The accuracy, usability, and touchless acquisition of state-of-the-art (SOTA) AFR systems have led to their ubiquitous adoption in a plethora of domains. 
However, this has also inadvertently sparked a community of
attackers that dedicate their time and effort to manipulate faces either physically~\cite{learning-deep-models-for-face-anti-spoofing-binary-or-auxiliary-supervision,on-disentangling-spoof-traces-for-generic-face-anti-spoofing} or digitally~\cite{dang2020manipulation}, in order to evade AFR systems~\cite{real_spoof}. 
AFR systems have been shown to be vulnerable to adversarial attacks resulting from perturbing an input probe~\cite{dong, gflm, advfaces, semantic_adv}. Even when the amount of perturbation is imperceptible to the human eye, such adversarial attacks can degrade the face recognition performance of SOTA AFR systems~\cite{advfaces}. With the growing dissemination of ``fake news" and ``deepfakes"~\cite{deep_fake}, research groups and social media platforms alike are pushing towards generalizable defense against continuously evolving adversarial attacks.\footnotetext{\url{https://bit.ly/2IkfSxk}}

\begin{figure}[t!]
    \centering
    \captionsetup{font=footnotesize}
    \scalebox{1.0}{
    \begin{minipage}{0.15\linewidth}
    \includegraphics[width=\linewidth]{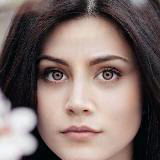}{\footnotesize \\ \centering0.21\par}
    \includegraphics[width=\linewidth]{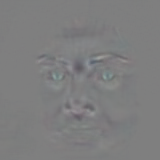}
    \centering {\footnotesize(a)~\cite{advfaces}}  
    \end{minipage}\;
    \begin{minipage}{0.15\linewidth}
    \includegraphics[width=\linewidth]{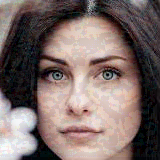}{\footnotesize \\ \centering0.27\par}
    \includegraphics[width=\linewidth]{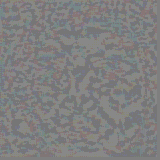}
    \centering {\footnotesize(b)~\cite{fgsm}}  
    \end{minipage}\;
    \begin{minipage}{0.15\linewidth}
    \includegraphics[width=\linewidth]{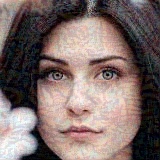}{\footnotesize \\ \centering0.28\par}
    \includegraphics[width=\linewidth]{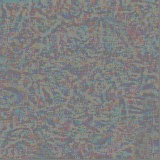} \centering {\footnotesize(b)~\cite{pgd}}  
    \end{minipage}\;
        \begin{minipage}{0.15\linewidth}
    \includegraphics[width=\linewidth]{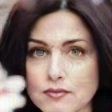}{\footnotesize \\ \centering0.32\par}
    \includegraphics[width=\linewidth]{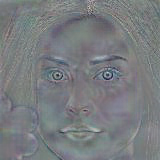}
    \centering {\footnotesize(c)~\cite{semantic_adv}}  
    \end{minipage}\;
        \begin{minipage}{0.15\linewidth}
    \includegraphics[width=\linewidth]{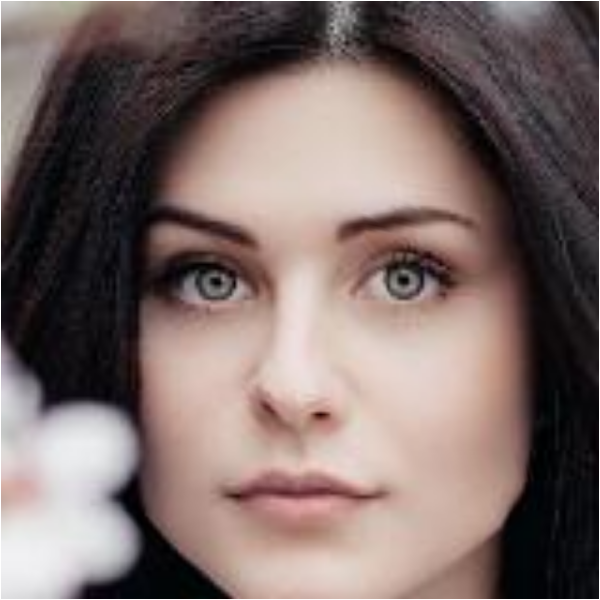}{\footnotesize \\ \centering0.34\par}
    \includegraphics[width=\linewidth]{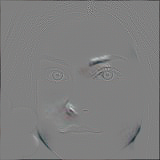}
    \centering {\footnotesize(d)~\cite{gflm}}  
    \end{minipage}\;
    \begin{minipage}{0.15\linewidth}
    \includegraphics[width=\linewidth]{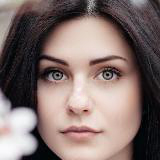}{\footnotesize \\ \centering0.35\par}
    \includegraphics[width=\linewidth]{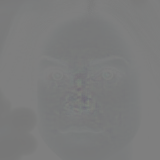}
    \centering {\footnotesize(e)~\cite{deepfool}} 
    \end{minipage}
    }\vspace{-2mm}
    \caption{\emph{(Top Row)} Adversarial faces synthesized via $6$ adversarial attacks used in our study. \emph{(Bottom Row)} Corresponding adversarial perturbations (gray indicates no change from the input). Notice the diversity in the perturbations. ArcFace scores between adversarial image and the unaltered gallery image (not shown here)  are given below each image. A score above $\textbf{0.36}$ indicates that two faces are of the same subject. Zoom in for details.}
    \label{fig:perturbation_sets}
\end{figure}

A considerable amount of research has focused on synthesizing adversarial attacks~\cite{fgsm, pgd, deepfool, advfaces, gflm, semantic_adv}. Obfuscation attempts (faces are perturbed such that they cannot be identified as the attacker) are more effective~\cite{advfaces}, computationally efficient to synthesize~\cite{fgsm, pgd}, and widely adopted~\cite{fawkes} compared to impersonation attacks (perturbed faces can automatically match to a target subject). Similar to prior defense efforts~\cite{massoli, smartbox}, this paper focuses on defending against obfuscation attacks (see Fig.~\ref{fig:frontpage}). Given an input probe image, $\bx$, an adversarial generator has two requirements under the obfuscation scenario: (1) synthesize an adversarial face image, $\bx_{adv} = \bx + \delta$, such that SOTA AFR systems fail to match $\bx_{adv}$ and $\bx$, and (2) limit the magnitude of perturbation $||\delta||_p$ such that $\bx_{adv}$ appears very similar to $\bx$ to humans.




A number of approaches have been proposed to defend against adversarial attacks. Their major shortcoming is~\emph{generalizability} to unseen adversarial attacks. Adversarial face perturbations may vary significantly (see Fig.~\ref{fig:perturbation_sets}). 
For instance, gradient-based attacks, such as FGSM~\cite{pgd} and PGD~\cite{pgd}, perturb every pixel in the face image, whereas, AdvFaces~\cite{advfaces} and SemanticAdv~\cite{semantic_adv} perturb only the salient facial regions, {\it e.g.}, eyes, nose, and mouth. On the other hand, GFLM~\cite{gflm} performs geometric warping to the face. Since the exact type of adversarial perturbation may not be known a priori, a defense system trained on a subset of adversarial attack types may have degraded performance on other unseen attacks. 

\begin{figure}[!t]
    \centering
    \captionsetup{font=footnotesize}
    \includegraphics[width=0.7\linewidth]{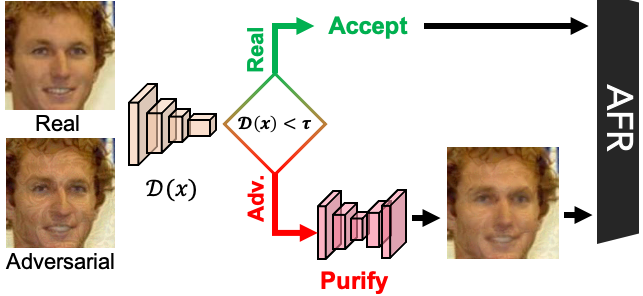}
    \vspace{-2mm}
    \caption{\emph{FaceGuard} employs a detector ($\ED$) to compute an adversarial score. Scores below detection threshold ($\tau$) passes the input to AFR, and high value invokes a purifier and sends the purified face to the AFR system.}
    \label{fig:rel_our}
\end{figure}

To the best of our knowledge, we take the first step towards a complete defense against adversarial faces by integrating an adversarial face generator, a detector, and a purifier into a unified framework, namely~\emph{FaceGuard} (see Fig.~\ref{fig:rel_our}). Robustness to unseen adversarial attacks is imparted via a stochastic generator that outputs diverse perturbations evading an AFR system, while a detector continuously learns to distinguish them from real faces. Concurrently, a purifier removes the adversarial perturbations from the synthesized image.

This work makes the following contributions: 
\setlist{topsep=0mm}
\begin{itemize}[noitemsep]
    \item A new self-supervised framework, namely~\emph{FaceGuard}, for defending against adversarial face images.~\emph{FaceGuard} combines benefits of adversarial training, detection, and purification into a unified defense mechanism trained in an end-to-end manner.
    \item With the proposed diversity loss, a generator is regularized to produce stochastic and challenging adversarial faces. We show that the diversity in output perturbations is sufficient for improving~\emph{FaceGuard}'s robustness to unseen attacks compared to utilizing pre-computed training samples from known attacks.
    \item Synthesized adversarial faces aid the detector to learn a tight decision boundary around real faces.~\emph{FaceGuard}'s detector achieves SOTA detection accuracies of $99.81\%$, $98.73\%$, and $99.35\%$ on $6$ unseen
    attacks on LFW~\cite{lfw}, Celeb-A~\cite{celeba}, and FFHQ~\cite{ffhq}.
    \item As the generator trains, a purifier concurrently removes perturbations from the synthesized  adversarial faces. With the proposed bonafide loss, the detector also guides purifier's training to ensure purified images are devoid of adversarial perturbations. At 0.1\% False Accept Rate,~\emph{FaceGuard}'s purifier enhances the True Accept Rate of ArcFace~\cite{arcface} from $34.27\%$ under no defense to $77.46\%$.
\end{itemize}


\Section{Related Work}

\begin{table*}[!t]
\scriptsize
\setlength{\tabcolsep}{2.2pt}
\captionsetup{font=footnotesize}
\centering
\begin{threeparttable}
\renewcommand{\arraystretch}{1.2}
\begin{tabularx}{\linewidth}{l| X X X X c}
\noalign{\hrule height 1.0pt}
\multicolumn{1}{l}{} & \textbf{Study} & \textbf{Method} & \textbf{Dataset} & \textbf{Attacks}  & \textbf{Self-Sup.}\\
\noalign{\hrule height 1.0pt}

\parbox[t]{2mm}{\multirow{7}{*}{\raisebox{5.5em}{\rotatebox[origin=c]{90}{Robustness}}}}
& Adv. Training~\cite{adv_train} (2017) & Train with adv. & ImageNet~\cite{imagenet} & FGSM~\cite{fgsm} & $\times$ \\
& RobGAN~\cite{robgan} (2019) & Train with generated adv. & CIFAR10~\cite{cifar}, ImageNet~\cite{imagenet} & PGD~\cite{pgd} & $\times$ \\
& Feat. Denoising~\cite{feat_denoising} (2019) & Custom network arch. & ImageNet~\cite{imagenet} & PGD~\cite{pgd} & $\times$ \\
& L2L~\cite{l2l} (2019) & Train with generated adv.  & MNIST~\cite{mnist}, CIFAR10~\cite{cifar} & FGSM~\cite{fgsm}, PGD~\cite{pgd}, C\&W~\cite{carlini} & \checkmark\\
\noalign{\hrule height 0.5pt}

\parbox[t]{2mm}{\multirow{17}{*}{\raisebox{13.0em}{\rotatebox[origin=c]{90}{Detection}}}}
& Gong~\etal~\cite{gong} (2017) & Binary CNN & MNIST~\cite{mnist}, CIFAR10~\cite{cifar} & FGSM~\cite{fgsm} & $\times$ \\
& UAP-D~\cite{uapd} (2018) & PCA+SVM & MEDS~\cite{meds}, MultiPIE~\cite{multipie}, PaSC~\cite{pasc} & UAP~\cite{uap} & $\times$\\
& SmartBox~\cite{smartbox} (2018) &  Adaptive Noise & Yale Face~\cite{yale} & DeepFool~\cite{deepfool}, EAD~\cite{ead}, FGSM~\cite{fgsm} & $\times$\\
& ODIN~\cite{odin} (2018) & Out-of-distribution Detection & CIFAR10~\cite{cifar}, ImageNet~\cite{imagenet} & OOD samples & $\times$\\
& Goswami~\etal~\cite{goswami2019detecting} (2019) & SVM on AFR Filters & MEDS~\cite{meds}, PaSC~\cite{pasc}, MBGC~\cite{mbgc} & Black-box, EAD~\cite{ead} & $\times$\\
& Steganalysis~\cite{steganalysis} (2019) & Steganlysis & ImageNet~\cite{imagenet} & FGSM~\cite{fgsm}, DeepFool~\cite{deepfool}, C\&W~\cite{carlini} & $\times$\\
& Massoli~\etal~\cite{massoli} (2020) & MLP/LSTM on AFR Filters & VGGFace2~\cite{vgg_face}  & BIM~\cite{bim}, FGSM~\cite{fgsm}, C\&W~\cite{carlini}  & $\times$\\
& Agarwal~\etal~\cite{agarwal_image_transform} (2020) & Image Transformation & ImageNet~\cite{imagenet}, MBGC~\cite{mbgc} & FGSM~\cite{fgsm}, PGD~\cite{pgd}, DeepFool~\cite{deepfool} & $\times$\\

\noalign{\hrule height 0.5pt}
\parbox[t]{2mm}{\multirow{7}{*}{\raisebox{4.0em}{\rotatebox[origin=c]{90}{Purification}}}}
& MagNet~\cite{magnet} (2017) & AE Purifier & MNIST~\cite{mnist}, CIFAR10~\cite{cifar} & FGSM~\cite{fgsm}, DeepFool~\cite{deepfool}, C\&W~\cite{carlini} & $\times$ \\
& DefenseGAN~\cite{defense_gan} (2018) & GAN & MNIST~\cite{mnist}, CIFAR10~\cite{cifar} & FGSM~\cite{fgsm}, C\&W~\cite{carlini} & $\times$ \\
& Feat. Distillation~\cite{feat_distillation} (2019) & JPEG-compression & MNIST~\cite{mnist}, CIFAR10~\cite{cifar} & FGSM~\cite{fgsm}, DeepFool~\cite{deepfool}, C\&W~\cite{carlini} & $\times$ \\
& NRP~\cite{self_supervised} (2020) & AE Purifier  & ImageNet~\cite{imagenet} & FGSM~\cite{fgsm} & $\checkmark$ \\
& A-VAE~\cite{avae} (2020) & Variational AE & LFW~\cite{lfw} & FGSM~\cite{fgsm}, PGD~\cite{pgd}, C\&W~\cite{carlini} & $\times$ \\

\noalign{\hrule height 1pt}
\multicolumn{1}{l}{} & \multirow{2}{*}{\emph{FaceGuard} (this study)} & \multirow{2}{*}{Adv. Generator + Detector  + Purifier} & \multirow{2}{*}{LFW~\cite{lfw}, Celeb-A~\cite{celeba}, FFHQ~\cite{ffhq}} & FGSM~\cite{fgsm}, PGD~\cite{pgd}, DeepFool~\cite{deepfool},  & \multirow{2}{*}{\checkmark} \\
\multicolumn{1}{l}{} & & & & \raisebox{0.2em}{AdvFaces~\cite{advfaces}, GFLM~\cite{gflm}, Semantic~\cite{semantic_adv}} & \\
\noalign{\hrule height 1pt}
\end{tabularx}
 \end{threeparttable}\vspace{-2mm}
\caption{Related work in adversarial defenses used as baselines in our study. Unlike majority of prior work,~\emph{FaceGuard} is self-supervised where no pre-computed adversarial examples are required for training.}
\label{tab:related_defense}
\end{table*}
\vspace{-3mm}
\Paragraph{Defense Strategies} \label{sec:related_defense}
In literature, a common defense strategy, namely~\emph{robustness} is to re-train the classifier we wish to defend with adversarial examples~\cite{fgsm, pgd, adv_train, l2l}. However, adversarial training has been shown to degrade classification accuracy on real (non-adversarial) images~\cite{robustness_cost1, robustness_cost2}.

In order to prevent degradation in AFR performance, a large number of adversarial defense mechanisms are deployed as a pre-processing step, namely~\emph{detection}, which involves training a binary classifier to distinguish between real and adversarial examples~\cite{stochastic, artifacts, gong, grosse, li_defense, hendrycks, guo, logit_pairing, metzen, cascade,xie, efficient,uapd, smartbox, massoli}. The attacks considered in
these studies~\cite{carlini_defense,obfuscated_gradients, magnet_fail, logit_fail} were initially proposed in the object recognition domain and they often fail to detect the attacks in a feature-extraction network setting, as in face recognition. Therefore, prevailing detectors against adversarial faces are demonstrated to be effective only in a highly constrained setting where the number of subjects is limited and fixed during training and testing~\cite{uapd, smartbox, massoli}.


Another pre-processing strategy, namely \emph{purification}, involves automatically removing adversarial perturbations in the input image prior to passing them to a face matcher~\cite{magnet, defense_gan, pixeldefend, self_supervised}. However, without a dedicated adversarial detector, these defenses may end up ``purifying" a real face image, resulting in high false reject rates.

In Tab.~\ref{tab:related_defense}, we summarize a few studies on adversarial defenses that are used as baselines in our work.



\Paragraph{Adversarial Attacks:} Numerous adversarial attacks have been proposed in literature~\cite{fgsm, pgd, advgan, papernot, carlini}. For example, Fast Gradient Sign Method (FGSM) generates an adversarial example by back-propagating through the target model~\cite{fgsm}. Other approaches optimize adversarial perturbation by
minimizing an objective function while satisfying certain constraints~\cite{pgd, deepfool, carlini}. We modify the objective functions of these attacks in order to craft adversarial faces that evade AFR systems. We evaluate~\emph{FaceGuard} on six unseen adversarial attacks that have high success rates in evading ArcFace~\cite{arcface}: FGSM~\cite{fgsm}, PGD~\cite{pgd}, DeepFool~\cite{deepfool}, AdvFaces~\cite{advfaces}, GFLM~\cite{gflm}, and SemanticAdv~\cite{semantic_adv} (see Tab.~\ref{tab:adv_faces}).

\Section{Limitations of State-of-the-Art Defenses}
\vspace{-3mm}\Paragraph{Robustness}
Adversarial training is regarded as one of the most effective defense method~\cite{fgsm, pgd, robgan} on small datasets including MNIST and CIFAR10.  
Whether this technique can scale to large datasets and a variety of different attack types (perturbation sets) has not yet been shown. Adversarial training is formulated as~\cite{fgsm, pgd}:
\begin{align}\label{eqn:adv_train}
  \min_{\theta}~\underset{(x,y)\sim\EP_{data}}{\IE}\left[\max_{\delta\in\Delta}\ell\left(f_{\theta}\left(x+\delta\right), y\right)\right],
\end{align}
where $(x,y)\sim\EP_{data}$ is the (image, label) joint distribution of data, $f_{\theta}\left(x\right)$ is the network parameterized by $\theta$, and $\ell\left(f_{\theta}\left(x\right), y\right)$ is the loss function (usually cross-entropy). Since the ground truth data distribution, $\EP_{data}$, is not known in practice, it is later replaced by the empirical distribution. Here, the network, $f_\theta$ is made robust by training with an adversarial noise ($\delta$) that maximally increases the classification loss. In other words, adversarial training involves training with the \emph{strongest} adversarial attack.


\begin{figure}[!t]
    \centering
    \captionsetup{font=footnotesize}
    \vspace{-2mm}
    \subfloat[Adversarial Training~\cite{adv_train}]{\includegraphics[width=0.5\linewidth]{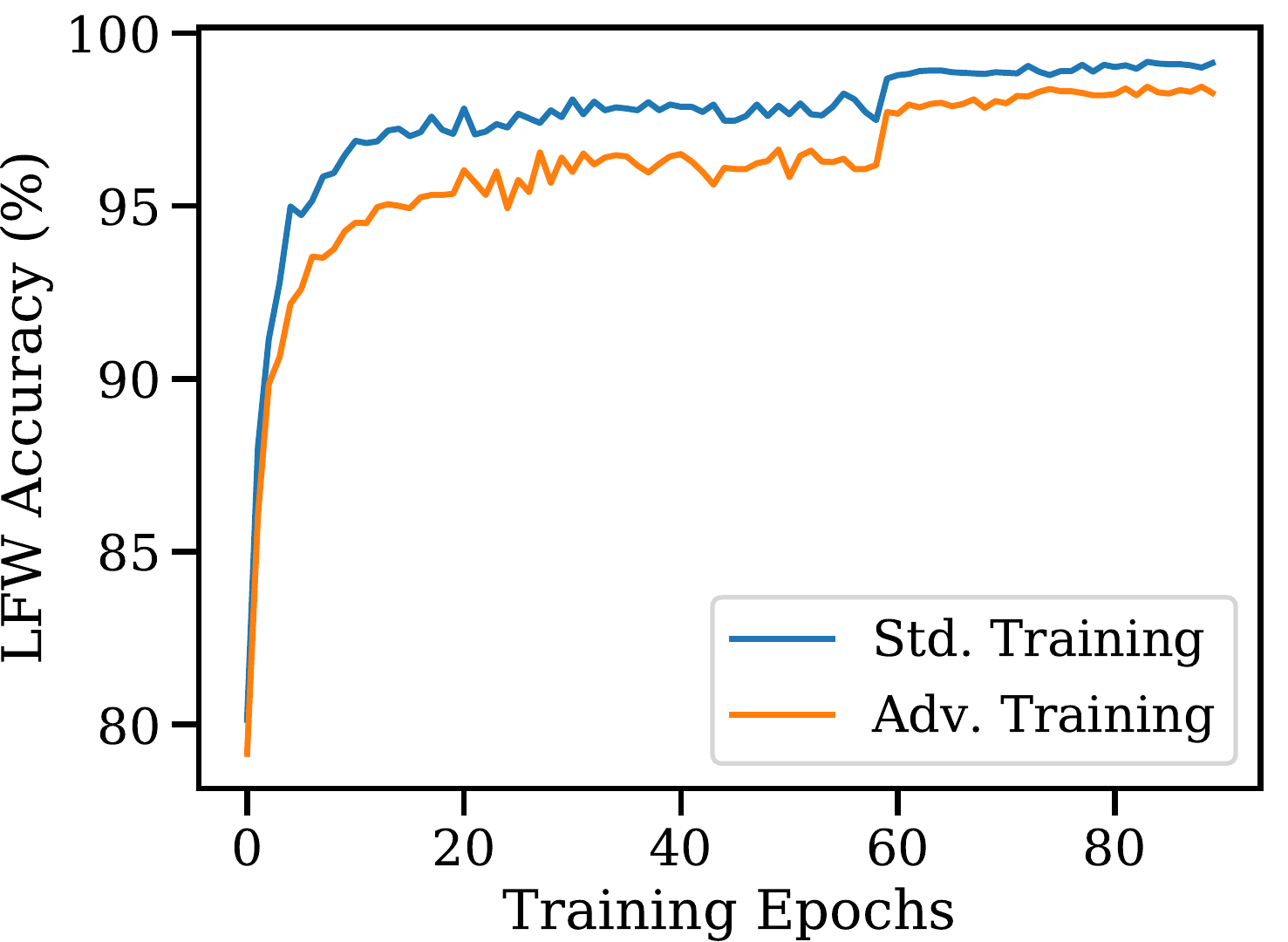}\label{fig:std_adv_lfw}}\hfill
    \subfloat[Detection~\cite{gong}]{\includegraphics[width=0.5\linewidth]{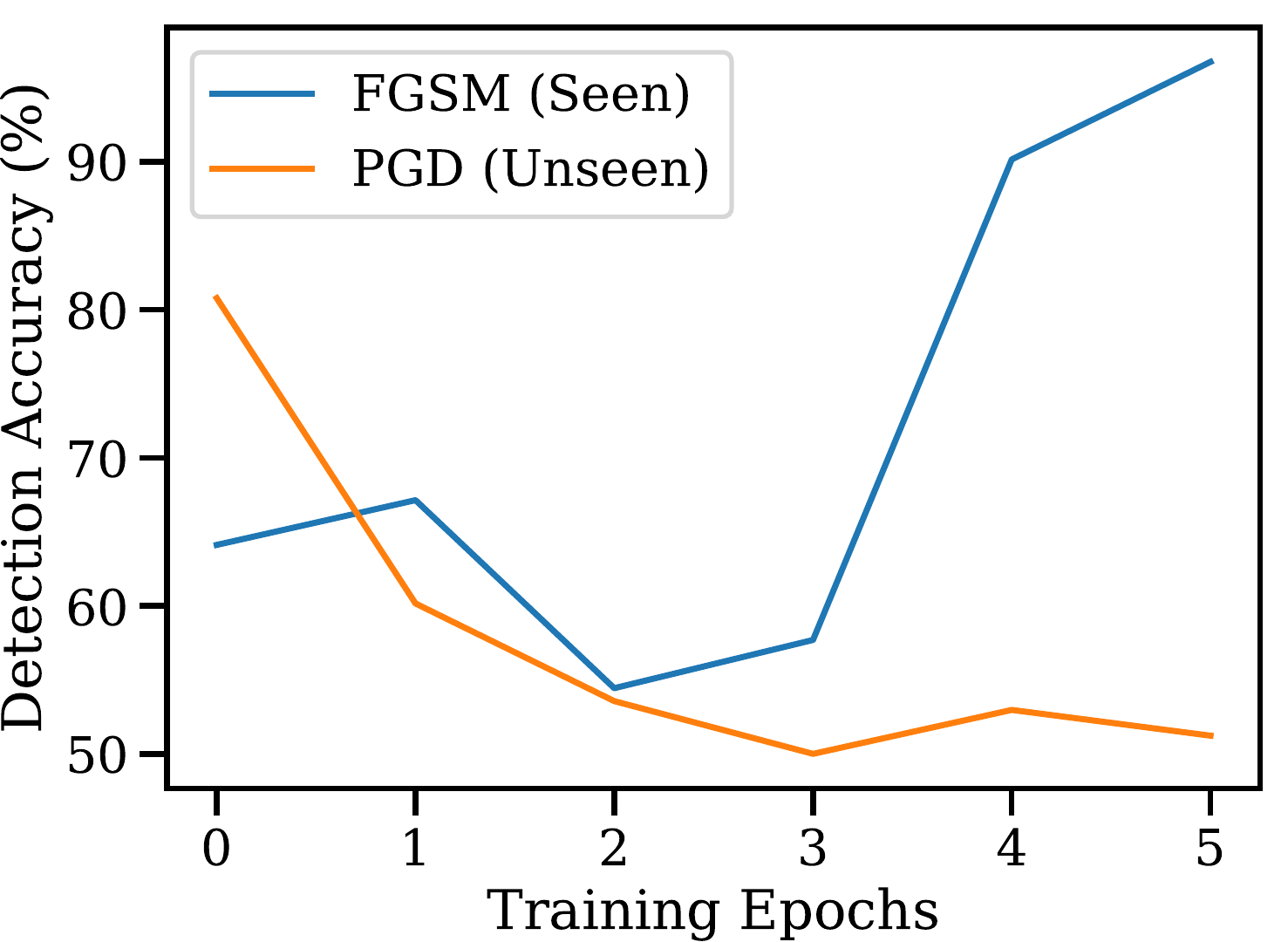}\label{fig:overfitting_detector}}
    \vspace{-2mm}
    \caption{(a) Adversarial training degrades AFR performance of FaceNet matcher~\cite{facenet} on real faces in LFW dataset compared to standard training. (b) A binary classifier trained to distinguish between real faces and FGSM~\cite{fgsm} attacks fails to detect unseen attack type, namely PGD~\cite{pgd}.}
    \label{fig:shortcomings}
\end{figure}

\begin{figure*}[!t]
    \centering
    \captionsetup{font=footnotesize}
    \includegraphics[width=0.8\linewidth]{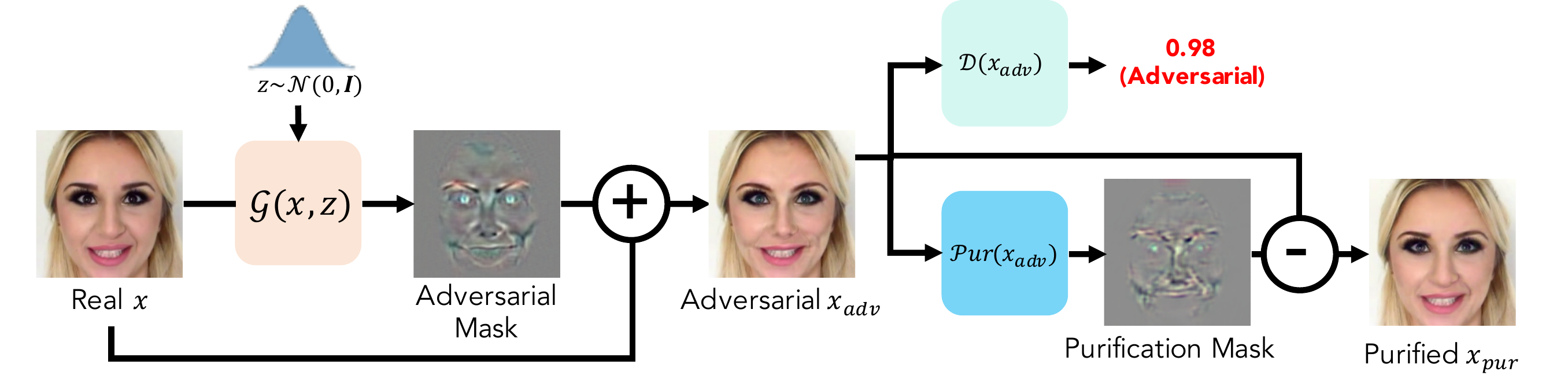}\vspace{-2mm}
    \caption{Overview of training the proposed~\emph{FaceGuard} in a self-supervised manner. An~\emph{adversarial generator}, $\EG$, continuously learns to synthesize challenging and diverse perturbations that evade a face matcher. At the same time, a~\emph{detector}, $\ED$, learns to distinguish between the synthesized adversarial faces and real face images. Perturbations residing in the synthesized adversarial faces are removed via a \emph{purifier}, $\EPur$.}
    \label{fig:overview}
\end{figure*}

The generalization of adversarial training has been in question~\cite{raghunathan, robgan, l2l, robustness_cost1, robustness_cost2}. It was shown that adversarial training can significantly reduce classification accuracy on real examples~\cite{robustness_cost1, robustness_cost2}. In the context of face recognition, we illustrate this by training two face matchers on CASIA-WebFace: (i) FaceNet~\cite{facenet} trained via the standard training process, and (ii) FaceNet~\cite{facenet} by adversarial training (FGSM\footnote{With max perturbation hyperparameter as $\epsilon = 8/256$.}). We then compute face recognition performance across training iterations on a separate testing dataset, LFW~\cite{lfw}. Fig.~\ref{fig:std_adv_lfw} shows that adversarial training drops the accuracy from $99.13\% \xrightarrow{} 98.27\%$. We gain the following insight: adversarial training may degrade AFR performance on real faces.




\Paragraph{Detection}\label{sec:adv_detect} ~Detection-based approaches employ a pre-processing step to ``detect" whether an input face is real or adversarial~\cite{gong, grosse, uapd, massoli}. A common approach is to utilize a binary classifier, $\mathcal{D}$, that maps a face image, $\bx \in \IR^{H\times W\times C}$ to $\{0, 1\}$, where $0$ indicates a real and $1$ an adversarial face. We train a binary classifier to distinguish between real and FGSM attack samples in CASIA-WebFace~\cite{casia}. In Fig.~\ref{fig:overfitting_detector}, we evaluate its detection accuracy on FGSM and PGD samples in LFW~\cite{lfw}.
We find that prevailing detection-based defense schemes may overfit to the specific adversarial attacks utilized for training.

\Section{FaceGuard}
Our defense aims to achieve robustness without sacrificing AFR performance on real face images. We posit that an adversarial defense trained alongside an adversarial generator in a~\emph{self-supervised} manner may improve robustness to unseen attacks. The main intuitions behind our defense mechanism are as follows:
\begin{itemize}[noitemsep,leftmargin=*]
    \item Since adversarial training may degrade AFR performance, we opt to obtain a robust adversarial \emph{detector} and~\emph{purifier} to detect and purify adversarial attacks.
    \item Given that prevailing detection-based methods tend to overfit to known adversarial perturbations (see Supp.), a detector and purifier trained on~\emph{diverse} synthesized adversarial perturbations may be more robust to unseen attacks.
    \item Sufficient diversity in synthesized perturbations can guide the detector to learn a tighter boundary around real faces. In this case, the detector itself can serve as a powerful supervision for the purifier.
    \item Lastly, pixels involved in the purification process may serve to  indicate adversarial regions in the input face.
\end{itemize}

\SubSection{Adversarial Generator}
The generalizability of an adversarial detector and purifier relies on the quality of the synthesized adversarial face images output by~\emph{FaceGuard}'s adversarial generator. We propose an adversarial generator that continuously learns to synthesize challenging and diverse adversarial face images.

The generator, denoted as $\EG$, takes an input real face image, $\bx \in \IR^{H\times W\times C}$, and outputs an adversarial perturbation $\EG(\bx,\bz)$, where $\bz \sim \mathcal{N}(0, \mathbf{I})$ is a random latent vector. Inspired by prevailing adversarial attack generators~\cite{fgsm, pgd, deepfool, advfaces, carlini}, we treat the output perturbation $\EG(\bx,\bz)$ as an additive \emph{perturbation mask}. The final adversarial face image, $\bx_{adv}$, is given by $\bx_{adv} = \bx + \EG(\bx,\bz)$.

In an effort to impart generalizability to the detector and purifier, we emphasize the following requirements of $\EG$:
\begin{itemize}[noitemsep]
    \item \textbf{Adversarial:} Perturbatation, $\EG(\bx,\bz)$, needs to be adversarial such that an AFR system cannot identify the adversarial face image $\bx_{adv}$ as the same person as the input probe $\bx$.
    \item \textbf{Visually Realistic:} Perturbation $\EG(\bx,\bz)$ should also be minimal such that $\bx_{adv}$ appears as a legitimate face image of the subject in the input probe $\bx$.
    \item \textbf{Stochastic:} For an input $\bx$, we require diverse adversarial perturbations, $\EG(\bx,\bz)$, for different latents $\bz$.
\end{itemize}


For satisfying all of the above requirements, we propose multiple loss functions to train the generator.


\noindent\textbf{Obfuscation Loss}  To ensure $\EG(\bx,\bz)$ is indeed \emph{adversarial}, we incorporate a white-box AFR system, $\EF$, to supervise the generator. Given an input face, $\bx$,  the generator aims to output an adversarial face, $\bx_{adv} = \bx + \EG(\bx,\bz)$ such that the face representations, $\EF(\bx)$ and $\EF(\bx_{adv})$, do not match. In other words, the goal is to minimize the cosine similarity between the two face representations\footnote{For brevity, we denote $\IE_\bx \equiv \IE_{\bx \in \mathcal{P}_{data}}$.}:
\begin{align}
    \EL_{obf} = \IE_{\bx}\left[\frac{\EF(\bx) \cdot \EF(\bx_{adv})}{\left|\left|\EF(\bx)\right|\right|\left|\left|\EF(\bx_{adv})\right|\right|}\right].
\end{align}

\noindent\textbf{Perturbation Loss}
With the identity loss alone, the generator may output perturbations with large magnitudes which will (a) be trivial for the detector to reject and (b) violate the visual realism requirement of $x_{adv}$. Therefore, we restrict the perturbations to be within $[-\epsilon, \epsilon]$ via a hinge loss:
\begin{align}
    \EL_{pt} = \IE_{\bx}\left[\max\left(\epsilon, \left|\left|\EG(\bx,\bz)\right|\right|_2\right)\right].
\end{align}

\noindent\textbf{Diversity Loss}
The above two losses jointly ensure that at each step, our generator learns to output challenging adversarial attacks. However, these attacks are deterministic; for an input image, we will obtain the same adversarial image. This may again lead to an inferior detector that overfits to a few deterministic perturbations seen during training. Motivated by studies of preventing mode collapse in GANs~\cite{mode_collapse}, we propose maximizing a diversity loss to promote stochastic perturbations per training iteration, $i$:
\begin{align}
    \EL_{div} = -\frac{1}{N_{ite}}\sum_{i=1}^{N_{ite}} \frac{\left|\left|\EG(\bx, \bz_1)^{(i)} - \EG(\bx, \bz_2)^{(i)}\right|\right|_1}{\left|\left|\bz_1 - \bz_2\right|\right|_1},
\end{align}
where $N_{ite}$ is the number of training iterations, $\EG(\bx, \bz)^{(i)}$ is the perturbation output at iteration $i$, and $(\bz_1,\bz_2)$ are two i.i.d.~samples from $\bz\sim \mathcal{N}(0,\mathbf{I})$. 
The diversity loss ensures that for two random latent vectors, $\bz_1$ and $\bz_2$, we will obtain two different perturbations $\EG(\bx, \bz_1)^{(i)}$ and $\EG(\bx, \bz_2)^{(i)}$.

\noindent\textbf{GAN Loss} Akin to prior work on GANs~\cite{goodfellow_gan, pix_to_pix}, we introduce a discriminator to encourage perceptual realism of the adversarial images. The discriminator, $Dsc$, aims to distinguish between probes, $\bx$, and synthesized faces $\bx_{adv}$ via a GAN loss:
\begin{align}
    \EL_{GAN} = \IE_{\bx}\left[{\log Dsc(\bx)}\right] +
   \IE_{\bx}{\left[\log(1-Dsc(\bx_{adv}))\right]}.
\end{align}

\SubSection{Adversarial Detector} 

Similar to prevailing adversarial detectors, the proposed detector also learns a decision boundary between real and adversarial images~\cite{grosse, gong, uapd, massoli}. A key difference, however, is that instead of utilizing pre-computed adversarial images from known attacks (\eg~FGSM and PGD) for training, the proposed detector learns to distinguish between real images and the~\emph{synthesized} set of diverse adversarial attacks output by the proposed adversarial generator in a self-supervised manner. This leads to the following advantage: \emph{our proposed framework does not require a large collection of pre-computed adversarial face images for training}.



We utilize a binary CNN for distinguishing between real input probes, $\bx$, and synthesized adversarial samples, $\bx_{adv}$. The detector is trained with the Binary Cross-Entropy loss:
\begin{align}
    \EL_{BCE} = \IE_{\bx}\left[\text{log}\ED(\bx)\right] + \IE_{\bx}\left[\text{log}\left(1 - \ED(\bx_{adv})\right)\right].
\end{align}



\SubSection{Adversarial Purifier}
The objective of the adversarial purifier is to recover the real face image $\bx$ given an adversarial face $\bx_{adv}$. We aim to automatically remove
the adversarial perturbations by training a neural network $\EPur$, referred as an adversarial purifier.

The adversarial purification process can be viewed as an inverted procedure of adversarial image synthesis. Contrary to the obfuscation loss in the adversarial generator, we require that the purified image, $\bx_{pur}$, successfully matches to the subject in the input probe $\bx$. Note that this can be achieved via a~\emph{feature recovery loss}, which is the opposite to the obfuscation loss, {\it i.e.}, $\EL_{fr}=-\EL_{obf}$.



Note that an adversarial face image, $\bx_{adv} = \bx + \delta$, is metrically close to the real image, $\bx$, in the input space. If we can estimate $\delta$, then we can retrieve the real face image. Here, the perturbations can be predicted by a neural network, $\EPur$. In other words, retrieving the purified image, $\bx_{pur}$ involves: (1) subtracting the perturbations from the adversarial image, $\bx_{pur} = \bx_{adv} - \EPur(\bx_{adv})$ and (2) ensuring that the~\emph{purification mask}, $\EPur(\bx_{adv})$, is small so that we do not alter the content of the face image by a large magnitude. Therefore, we propose a hybrid perceptual loss that (1) ensures $\bx_{pur}$ is as close as possible to the real image, $\bx$ via a $\ell_{1}$ reconstruction loss and (2) a loss that minimizes the amount of alteration, $\EPur(\bx_{adv})$:
\begin{align}
\begin{split}
    \EL_{perc} = \IE_{\bx}\left|\left|\bx_{pur} - \bx \right|\right|_1 + \left|\left|\EPur(\bx_{adv})\right|\right|_2.
\end{split}
\end{align}

Finally, we also incorporate our detector to guide the training of our purifier. Note that, due to the diversity in synthesized adversarial faces, the proposed detector learns a tight decision boundary around real faces. This can serve as a strong self-supervisory signal to the purifier for ensuring that the purified images belong to the real face distribution. Therefore, we also incorporate the detector as a discriminator for the purifier via the proposed bonafide loss:
\begin{align}
    \EL_{bf} = \IE_{\bx}\left[\text{log} \ED(\bx_{pur})\right].
\end{align}

\begin{table}[!t]
\centering
\ra{1.0}
\footnotesize
\captionsetup{font=footnotesize}
\begin{threeparttable}
\begin{tabular}{lMM}
\toprule \textbf{Attacks} & \text{TAR (\%) @ $0.1\%$ FAR} (\downarrow) & \text{SSIM} (\uparrow)\\
\midrule
FGSM~\cite{fgsm} & 26.23 & 0.83 \pm 0.24\\
PGD~\cite{pgd} & 04.91 & 0.89 \pm 0.12\\
DeepFool~\cite{deepfool} & 36.18 & 0.91 \pm 0.09\\
AdvFaces~\cite{advfaces} & 00.17 & 0.89 \pm 0.02\\
GFLM~\cite{gflm} & 68.03 & 0.55 \pm 0.14\\
SemanticAdv~\cite{semantic_adv} & 70.05 & 0.71 \pm 0.21\\
\midrule
No Attack & 99.82 & 1.00 \pm 0.00\\
\bottomrule
\end{tabular}
\end{threeparttable}\vspace{-2mm}
\caption{Face recognition performance of ArcFace~\cite{arcface} under adversarial attack and average structural similarities (SSIM) between probe and adversarial images for obfuscation attacks on $485K$ genuine pairs in LFW~\cite{lfw}.}
\label{tab:adv_faces}
\end{table}

\begin{table*}[!t]
\centering
\ra{1.0}
\footnotesize
\captionsetup{font=footnotesize}
\begin{threeparttable}
\begin{tabular}{l|lMMMMMMMM}
\toprule \multicolumn{1}{l}{} & \textbf{Detection Accuracy (\%)} & \textbf{Year} & \textbf{FGSM~\cite{fgsm}} & \textbf{PGD~\cite{pgd}} & \textbf{DpFl.~\cite{deepfool}} & \textbf{AdvF.~\cite{advfaces}} & \textbf{GFLM~\cite{gflm}} & \textbf{Smnt.~\cite{semantic_adv}} & \textbf{Mean $\pm$ Std.}\\
\midrule
\parbox[t]{2mm}{\multirow{7}{*}{\raisebox{5.5em}{\rotatebox[origin=c]{90}{General}}}}
& Gong~\etal\cite{gong} & 2017 &  98.94 & 97.91 & 95.87 & 92.69 & \mathbf{99.92} & \mathbf{99.92} & 97.54 \pm 02.82\\
& ODIN~\cite{odin} & 2018 & 83.12   &  84.39 & 71.74 & 50.01 & 87.25 & 85.68 & 77.03 \pm 14.34 \\
& Steganalysis~\cite{steganalysis} & 2019 & 88.76 & 89.34 & 75.97 & 54.30 & 58.99 & 78.62 & 74.33 \pm 14.77\\
\midrule
\parbox[t]{2mm}{\multirow{11}{*}{\raisebox{5.5em}{\rotatebox[origin=c]{90}{Face}}}}
& UAP-D~\cite{uapd} & 2018 & 61.32 & 74.33 & 56.78 & 51.11 & 65.33 & 76.78 & 64.28 \pm 09.97\\
& SmartBox~\cite{smartbox} & 2018 & 58.79 & 62.53 & 51.32 & 54.87 & 50.97 & 62.14 & 56.77 \pm 05.16\\
& Goswami~\etal\cite{goswami2019detecting} & 2019 & 84.56 & 91.32  & 89.75 & 76.51 & 52.97 & 81.12 & 79.37 \pm 14.04\\
& Massoli~\etal\cite{massoli}~(MLP) & 2020 & 63.58 & 76.28 & 81.78 & 88.38 & 51.97 & 52.98 & 69.16 \pm 15.29\\
& Massoli~\etal\cite{massoli}~(LSTM) & 2020 & 71.53 & 76.43 & 88.32 & 75.43 & 53.76 & 55.22 & 70.11 \pm 13.35\\
& Agarwal~\etal\cite{agarwal_image_transform} & 2020 & 94.44 & 95.38  & 91.19 & 74.32 & 51.68 & 87.03 & 87.03 \pm 16.86\\
\midrule
\multicolumn{1}{l}{} & \emph{Proposed FaceGuard} & 2021 & \mathbf{99.85} & \mathbf{99.85} & \mathbf{99.85} & \mathbf{99.84} & 99.61 & 99.85 & \mathbf{99.81 \pm 00.10}\\
\bottomrule
\end{tabular}
\end{threeparttable}\vspace{-2mm}
\caption{Detection accuracy of SOTA adversarial face detectors in classifying six adversarial attacks synthesized for the LFW dataset~\cite{lfw}. Detection threshold is set as $0.5$ for all methods. All baseline methods require training on pre-computed adversarial attacks on CASIA-WebFace~\cite{casia}. On the other hand, the proposed~\emph{FaceGuard} is self-guided and generates adversarial attacks on the fly. Hence, it can be regarded as a~\emph{black-box} defense system.}
\label{tab:detection}
\end{table*}
\SubSection{Training Framework}
We train the entire~\emph{FaceGuard} framework in Fig.~\ref{fig:overview} in an end-to-end manner with the following objectives:
\begin{align*}
    \begin{split}
    &\min_{\EG}\EL_{\EG} = \EL_{GAN} + \lambda_{obf}\cdot\EL_{obf}  + \lambda_{pt}\cdot\EL_{pt} - \lambda_{div}\cdot\EL_{div},
    \end{split}\\
    &\min_{\ED}\EL_{\ED} = \EL_{BCE},\\
    \begin{split}
    &\min_{\EPur}\EL_{\EPur} = \lambda_{fr}\cdot\EL_{fr}  + \lambda_{perc}\cdot\EL_{perc} + \lambda_{bf}\cdot\EL_{bf}.
    \end{split}
\end{align*}
At each training iteration, the generator attempts to fool the discriminator by synthesizing visually realistic adversarial faces while the discriminator learns to distinguish between real and synthesized images. On the other hand, in the same iteration, an external critic network, namely detector $\ED$, learns a decision boundary between real and synthesized adversarial samples. Concurrently, the purifier $\EPur$ learns to invert the adversarial synthesis process. Note that there is a key difference between the discriminator and the detector: the generator is designed to specifically~\emph{fool} the discriminator but not necessarily the detector. We will show in our experiments that this crucial step prevents the detector from predicting $\ED(\bx) = 0.5$ for all $\bx$ (see Tab.~\ref{tab:analysis}).

\Section{Experimental Results}
\SubSection{Experimental Settings}
\Paragraph{Datasets} We train~\emph{FaceGuard} on real face images in CASIA-WebFace~\cite{casia} dataset and then evaluate on real and adversarial faces synthesized for LFW~\cite{lfw}, Celeb-A~\cite{celeba} and FFHQ~\cite{ffhq} datasets.
 CASIA-WebFace~\cite{casia} comprises of $494,414$ face images from $10,575$\footnote{We removed $84$ subjects in CASIA-WebFace that overlap with LFW.} different subjects.
LFW~\cite{lfw} contains $13,233$ face images of $5,749$ subjects. Since we evaluate defenses under obfuscation attacks, we consider subjects with at least two face images\footnote{Obfuscation attempts only affect genuine pairs (two face images pertaining to the same subject).}. After this filtering, $9,164$ face images of $1,680$ subjects in LFW are available for evaluation. For brevity, experiments on CelebA and FFHQ are provided in Supp.

\Paragraph{Implementation} The adversarial generator and purifier employ a convolutional encoder-decoder. The latent variable $\mathbf{z}$, a $128$-dimensional feature vector, is fed as input to the generator through spatial padding and concatenation. The adversarial detector, a $4$-layer binary CNN, is trained jointly with the generator and purifier. Empirically, we set $\lambda_{obf} = \lambda_{fr} = 10.0$, $\lambda_{pt} = \lambda_{perc} = 1.0$, $\lambda_{div} = 1.0$, $\lambda_{bf} = 1.0$ and $\epsilon = 3.0$. Training and network architecture details are provided in Supp.

\Paragraph{Face Recognition Systems} In this study, we use two AFR systems: FaceNet~\cite{facenet} and ArcFace~\cite{arcface}. Recall that the proposed defense utilizes a face matcher, $\EF$, for guiding the training process of the generator. However, the deployed AFR system may not be known to the defense system a priori. Therefore, unlike prevailing defense mechanisms~\cite{uapd,smartbox,massoli}, we evaluate the effectiveness of the proposed defense on an AFR system~\emph{different} from $\EF$. We highlight the effectiveness of our proposed defense:~\emph{FaceGuard is trained on FaceNet, while the adversarial attack test set is designed to evade ArcFace.} Obfuscation attempts perturb real probes into adversarial ones. Ideally, deployed AFR systems (say, ArcFace), should be able to match a genuine pair comprised of an adversarial probe and a real enrolled face of the same subject. Therefore, regardless of real or adversarial probe, we assume that genuine pairs should~\emph{always} match as ground truth. Tab.~\ref{tab:adv_faces} provides AFR performance of ArcFace under $6$ SOTA adversarial attacks for $484,514$ genuine pairs in LFW. It appears that some attacks, \eg,~AdvFaces~\cite{advfaces}, are effective in both low TAR and high SSIM, while some are less capable in both metrics.




\SubSection{Comparison with State-of-the-Art Defenses}
In this section, we compare the proposed~\emph{FaceGuard} to prevailing defenses. We evaluate all methods via publicly available repositories provided by the authors (see Supp.). All baselines are trained on CASIA-WebFace~\cite{casia}.

\Paragraph{SOTA Detectors} Our baselines include $9$ SOTA detectors proposed both for general objects~\cite{gong, odin, steganalysis} and adversarial faces~\cite{uapd, goswami2019detecting, smartbox, massoli, agarwal_image_transform}. The detectors are trained on real and adversarial faces images synthesized via six adversarial generators for CASIA-WebFace~\cite{casia}. Unlike all the baselines,~\emph{FaceGuard}'s detector does not utilize any pre-computed adversarial attack for training. We compute the classification accuracy for all methods on a dataset comprising of $9,164$ real images and $9,164$ adversarial face images per attack type in LFW.

In Tab.~\ref{tab:detection}, we find that compared to the baselines,~\emph{FaceGuard} achieves the highest detection accuracy. Even when the $6$ adversarial attack types are encountered in training, a binary CNN~\cite{gong}, still falls short compared to~\emph{FaceGuard}.  This is likely because~\emph{FaceGuard} is trained on a  diverse set of adversarial faces from the proposed generator. While the binary CNN has a small drop compared to FaceGuard in the seen attacks ($99.81\% \xrightarrow{} 97.54\%$), it drops significantly on unseen adversarial attacks in testing (see Supp.).

\begin{figure}[!t]
    \centering
    \captionsetup{font=footnotesize}
    \subfloat[Real faces falsely detected as adversarial]{
    \begin{minipage}{0.23\linewidth}
    \includegraphics[width=\linewidth]{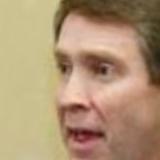}\\
    \centering {\footnotesize $0.77$}
    \end{minipage}\;
    \begin{minipage}{0.23\linewidth}
    \includegraphics[width=\linewidth]{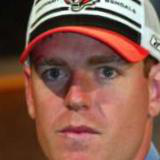}\\
    \centering {\footnotesize $0.67$}
    \end{minipage}\;
    \begin{minipage}{0.23\linewidth}
    \includegraphics[width=\linewidth]{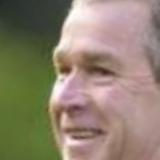}\\
    \centering {\footnotesize $0.96$}
    \end{minipage}\;
    \begin{minipage}{0.23\linewidth}
    \includegraphics[width=\linewidth]{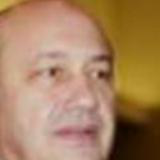}\\
    \centering {\footnotesize $0.99$}
    \end{minipage}\;
    }\\\vspace{-3mm}
    \subfloat[Adversarial faces falsely detected as real]{
    \begin{minipage}{0.23\linewidth}
    \includegraphics[width=\linewidth]{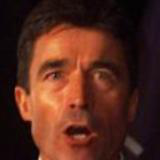}\\
    \centering {\footnotesize Real}
    \end{minipage}\;
    \begin{minipage}{0.23\linewidth}
    \includegraphics[width=\linewidth]{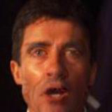}\\
    \centering {\footnotesize AdvFaces ($0.42$)}
    \end{minipage}\;
    \begin{minipage}{0.23\linewidth}
    \includegraphics[width=\linewidth]{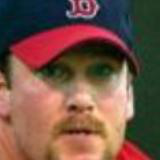}\\
    \centering {\footnotesize Real}
    \end{minipage}\;
    \begin{minipage}{0.23\linewidth}
    \includegraphics[width=\linewidth]{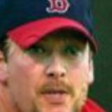}\\
    \centering {\footnotesize AdvFaces ($0.28$)}
    \end{minipage}\;
    }\vspace{-2mm}
    \caption{Examples where the proposed~\emph{FaceGuard} fails to correctly detect (a) real faces and (b) adversarial faces. Detection scores $\in [0,1]$ are given below each image, where $0$ indicates real and $1$ indicates adversarial face.}
    \label{fig:failures}
\end{figure}

Compared to hand-crafted features, such as PCA+SVM in UAP-D~\cite{uapd} and entropy detection in SmartBox~\cite{smartbox},~\emph{FaceGuard} achieves superior detection results. Some baselines utilize AFR features for identifying adversarial inputs~\cite{massoli, goswami2019detecting}. We find that intermediate AFR features primarily represent the identity of the input face and do not appear to contain highly discriminative information for detecting adversarial faces. 

Despite the robustness,~\emph{FaceGuard} misclassifies $28$ out of $9,164$ real images in LFW~\cite{lfw} and falsely predicts $46$ out of $54,984$ adversarial faces as real. From the latter, $44$ are warped faces via GFLM~\cite{gflm} and the remaining two are synthesized via AdvFaces~\cite{advfaces}. We find that~\emph{FaceGuard} tends to misclassify real faces under extreme poses and adversarial faces that are occluded (\eg, hats) (see Fig.~\ref{fig:failures}).

\Paragraph{Comparison with Adversarial Training \& Purifiers}
We also compare with prevailing defenses designing robust face matchers~\cite{adv_train,robgan,l2l} and purifiers~\cite{self_supervised, magnet, defense_gan}. We conduct a verification experiment by considering all possible genuine pairs (two faces belonging to the same subject) in LFW~\cite{lfw}. For one probe in a genuine pair, we craft six different adversarial probes (one per attack type). In total, there are $484,514$ real pairs and $\sim3M$ adversarial pairs. For a fixed match threshold\footnote{We compute the threshold at 0.1\% FAR on all possible image pairs in LFW, \eg, threshold @ 0.1\% FAR for ArcFace is set at $0.36$.}, we compute the True Accept Rate (TAR) of successfully matching two images in a real or adversarial pair in Tab.~\ref{tab:fr_perf}. 
In other words, TAR is defined here as the ratio of genuine pairs above the match threshold.

ArcFace without any adversarial defense system achieves $34.27\%$ TAR at $0.1\%$ FAR under attack. Adversarial    training~\cite{adv_train, robgan, l2l} inhibits the feature space of ArcFace, resulting in worse performance on both real and adversarial pairs. On the other hand, purification methods~\cite{magnet,defense_gan,self_supervised} can better retain face features in real pairs but their performance under attack is still undesirable.  

Instead, the proposed~\emph{FaceGuard} defense system first detects whether an input face image is real or adversarial. If input faces are adversarial, they are further purified. From Tab.~\ref{tab:fr_perf}, we find that our defense system significantly outperforms SOTA baselines in protecting ArcFace~\cite{arcface} against attacks. Specifically,~\emph{FaceGuard}'s purifier enhances ArcFace's average TAR at $0.1\%$ FAR under all six attacks (see Tab.~\ref{tab:adv_faces}) from $34.27\% \xrightarrow{} 77.46\%$. In addition,~\emph{FaceGuard} also maintains similar face recognition performance on real faces (TAR on real pairs drop from $99.82\% \xrightarrow{} 99.81\%$).
Therefore, our proposed defense system ensures that benign users will not be incorrectly rejected while malicious attempts to evade the AFR system will be curbed.

\begin{table}
\centering
\ra{1.0}
\footnotesize
\captionsetup{font=footnotesize}
\begin{threeparttable}
\begin{tabular}{lccMM}
\toprule \textbf{Defenses} & \textbf{Year} & \textbf{Strategy} &  \multirow[t]{2}{*}{\textbf{Real}} & \multirow[t]{2}{*}{\textbf{Attacks}}\\
& & 485K~\text{pairs} & 3M~\text{pairs}\\
\midrule
No-Defense & $-$ & - & 99.82 & 34.27 \\ \midrule
Adv. Training~\cite{adv_train} & 2017 & Robustness & 96.42 & 11.23\\
Rob-GAN~\cite{robgan} & 2019 & Robustness & 91.35 & 13.89\\
Feat. Denoising~\cite{feat_denoising} & 2019 & Robustness & 87.61 & 17.97 \\
L2L~\cite{l2l} & 2019 & Robustness & 96.89 & 16.76\\ \midrule 
MagNet~\cite{magnet} & 2017 & Purification & 94.47 & 38.32\\ 
DefenseGAN~\cite{defense_gan} & 2018 & Purification & 96.78 & 39.21\\
Feat. Distillation~\cite{feat_distillation} & 2019 & Purification & 94.64 & 41.77\\
NRP~\cite{self_supervised} & 2020 & Purification & 97.54 & 61.44\\
A-VAE~\cite{avae} & 2020 & Purification & 93.71 & 51.99 \\ \midrule
\emph{Proposed FaceGuard} & 2021 &  Purification & \mathbf{99.81} & \mathbf{77.46}\\
\bottomrule
\end{tabular}
\end{threeparttable}\vspace{-2mm}
\caption{AFR performance (TAR (\%) @ 0.1\% FAR) of ArcFace under no defense and when ArcFace is trained via SOTA robustness techniques~\cite{adv_train,robgan,l2l} or SOTA purifiers~\cite{magnet, defense_gan}. \emph{FaceGuard} correctly passes majority of real faces to ArcFace and also purifies adversarial attacks.}
\label{tab:fr_perf}
\end{table}
\SubSection{Analysis of Our Approach}

\begin{table}[!t]
\centering
\ra{1.0}
\footnotesize
\captionsetup{font=footnotesize}
\begin{threeparttable}
\begin{tabular}{c|l|M|M}
\toprule & \textbf{Model} & \textbf{AdvFaces~\cite{advfaces}} & \textbf{Mean $\pm$ Std.}\\
\midrule
\parbox[t]{2mm}{\multirow{3}{*}{\rotatebox[origin=c]{90}{Gen. $\EG$}}} & Without $\EG$ & 91.72 & 97.12 \pm 04.54 \\
& Without $\EL_{div}$ & 95.42 & 98.23 \pm 01.33\\
& \emph{With $\EG$ and $\EL_{div}$} & \mathbf{99.84} & \mathbf{99.81 \pm 00.10}\\
\midrule
\parbox[t]{2mm}{\multirow{3}{*}{\rotatebox[origin=c]{90}{Det. $\ED$}}} & $\ED$ as Discriminator & 50.00 & 75.25 \pm 21.19\\
& $\ED$ via Pre-Computed $\EG$  & 52.01 & 69.37 \pm 19.91\\
& \emph{$\ED$ as Online Detector} & \mathbf{99.84} & \mathbf{99.81 \pm 00.10}\\
\bottomrule
\end{tabular}
\end{threeparttable}\vspace{-2mm}
\caption{Ablating training schemes of the generator $\EG$ and detector $\ED$. All models are trained on CASIA-WebFace~\cite{casia}. \emph{(Col. 3)} We compute the detection accuracy in classifying real faces in LFW~\cite{lfw} and the most challenging adversarial attack in Tab.~\ref{tab:adv_faces}, AdvFaces~\cite{advfaces}. \emph{(Col. 4)} The avg. and std. dev. of detection accuracy across all 6 adversarial attacks.}
\label{tab:analysis}
\end{table}


\begin{figure}[!t]
    \centering
    \captionsetup{font=footnotesize}
    \subfloat[Adversarial faces via random latents within the same iteration.]{
    \begin{minipage}{0.23\linewidth}
    \includegraphics[width=\linewidth]{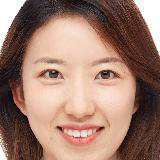}\\[0.1em]
    \includegraphics[width=\linewidth]{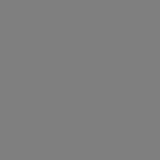}
    \centering {\footnotesize Input Probe ($\bx$)}
    \end{minipage}\;
    \begin{minipage}{0.23\linewidth}
    \includegraphics[width=\linewidth]{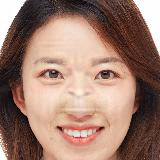}\\[0.1em]
    \includegraphics[width=\linewidth]{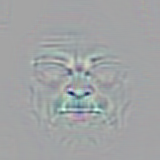}
    \centering {\footnotesize  $\EG(\bx,\bz_1)$}
    \end{minipage}\;
    \begin{minipage}{0.23\linewidth}
    \includegraphics[width=\linewidth]{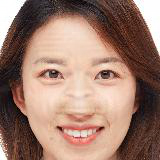}\\[0.1em]
    \includegraphics[width=\linewidth]{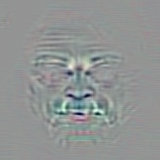}
    \centering {\footnotesize $\EG(\bx,\bz_2)$}
    \end{minipage}\;
      \begin{minipage}{0.23\linewidth}
    \includegraphics[width=\linewidth]{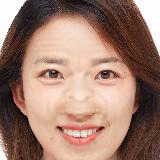}\\[0.1em]
    \includegraphics[width=\linewidth]{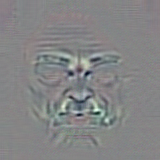}
    \centering {\footnotesize $\EG(\bx,\bz_3)$}
    \end{minipage}\;}\\\vspace{-2mm}
    \subfloat[Adversarial faces at different training iterations.]{
    \begin{minipage}{0.23\linewidth}
    \includegraphics[width=\linewidth]{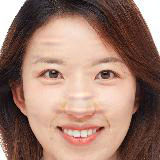}\\[0.1em]
    \includegraphics[width=\linewidth]{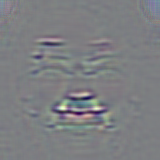}
    \centering {\footnotesize Iteration: $5K$}
    \end{minipage}\;
    \begin{minipage}{0.23\linewidth}
    \includegraphics[width=\linewidth]{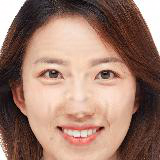}\\[0.1em]
    \includegraphics[width=\linewidth]{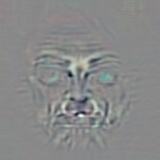}
    \centering {\footnotesize Iteration: $20K$}
    \end{minipage}\;
      \begin{minipage}{0.23\linewidth}
    \includegraphics[width=\linewidth]{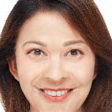}\\[0.1em]
    \includegraphics[width=\linewidth]{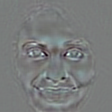}
    \centering {\footnotesize Iteration: $60K$}
    \end{minipage}\;
      \begin{minipage}{0.23\linewidth}
    \includegraphics[width=\linewidth]{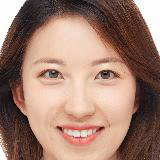}\\[0.1em]
    \includegraphics[width=\linewidth]{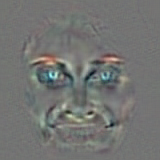}
    \centering {\footnotesize Iteration: $100K$}
    \end{minipage}\;
    }\vspace{-2mm}
    \caption{Adversarial faces synthesized by~\emph{FaceGuard} during training. Note the diversity in  perturbations (a) within and (b) across iterations.}
    \label{fig:diverse}
\end{figure}

\Paragraph{Quality of the Adversarial Generator}
In Tab.~\ref{tab:analysis}, we see that without the proposed adversarial generator (``Without $\EG$"),~\ie, a detector trained on the six known attack types, suffers from high standard deviation. Instead, training a detector with a deterministic $\EG$ (``Without $\EL_{div}$"), leads to better generalization across attack types, since the detector still encounters variations in synthesized images as the generator learns to better generate adversarial faces. However, such a detector is still prone to overfitting to a few deterministic perturbations output by $\EG$. Finally,~\emph{FaceGuard} with the diversity loss introduces diverse perturbations within and across training iterations (see Fig.~\ref{fig:diverse}).


\begin{figure}[!t]
\captionsetup{font=footnotesize}
\footnotesize
\centering\begin{tabular}{c@{ }c@{ }c@{ }c@{ }c@{ }}
\textbf{Probe} & \textbf{AdvFaces~\cite{advfaces}} & \textbf{Localization} & \textbf{Purified} \\
\includegraphics[width=.24\textwidth]{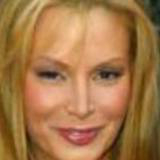}&
\includegraphics[width=.24\textwidth]{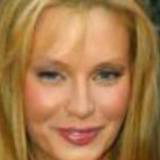}&
\includegraphics[width=.24\textwidth]{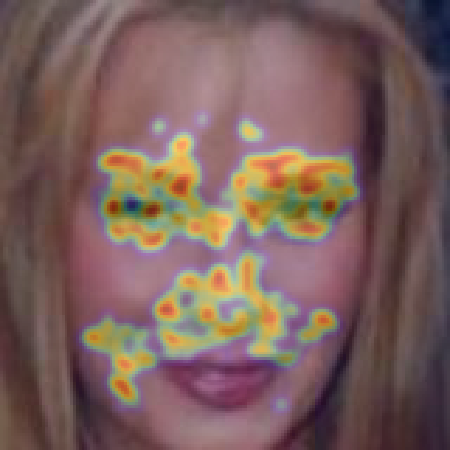}&
\includegraphics[width=.24\textwidth]{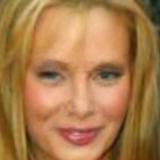}\\
\textbf{ArcFace/SSIM:} & $-0.30/0.89$ & & \footnotesize $0.62/0.91$\\
\end{tabular}\vspace{-2mm}
\caption{\emph{FaceGuard} successfully purifies the adversarial image (red regions indicate adversarial perturbations localized by our purification mask). ArcFace~\cite{arcface} scores $\in [-1,1]$ and SSIM $\in [0,1]$ between an adversarial/purified probe and input probe are given below each image.} 
\label{fig:qual_results}
\end{figure}

\Paragraph{Quality of the Adversarial Detector} The discriminator's task is similar to the detector; determine whether an input image is real or fake/adversarial. The key difference is that the generator is enforced to fool the discriminator, but not the detector. If we replace the discriminator with an adversarial detector, the generator continuously attempts to fool the detector by synthesizing images that are as close as possible to the real image distribution. By design, such a detector should converge to $Disc(\bx) = 0.5$ for all $\bx$ (real or adversarial). As we expect, in Tab.~\ref{tab:analysis}, we cannot rely on predictions made by such a detector (``$\ED$ as Discriminator"). We try another variant: we first train the generator $\EG$ and then train a detector to distinguish between real and pre-computed attacks via $\EG$ (``$\ED$ via Pre-Computed $\EG$"). As we expect, the proposed methodology of training the detector in an online fashion by utilizing the synthesized adversarial samples output by $\EG$ at any given iteration leads to a significantly robust detector (``$\ED$ as Online Detector"). This can likely be attributed to the fact that a detector trained on-line encounters a much larger variation as the generator trains alongside. ``$\ED$ via Pre-Computed $\EG$" is exposed only to within-iteration variations (from random latent sampling), however, `$\ED$ as Online Detector" encounters variations~\emph{both} within and across training iterations (see Fig.~\ref{fig:diverse}).

\begin{figure}[!t]
\centering
\captionsetup{font=footnotesize}
\footnotesize\vspace{-3mm}
\subfloat[]{\includegraphics[width=0.5\linewidth]{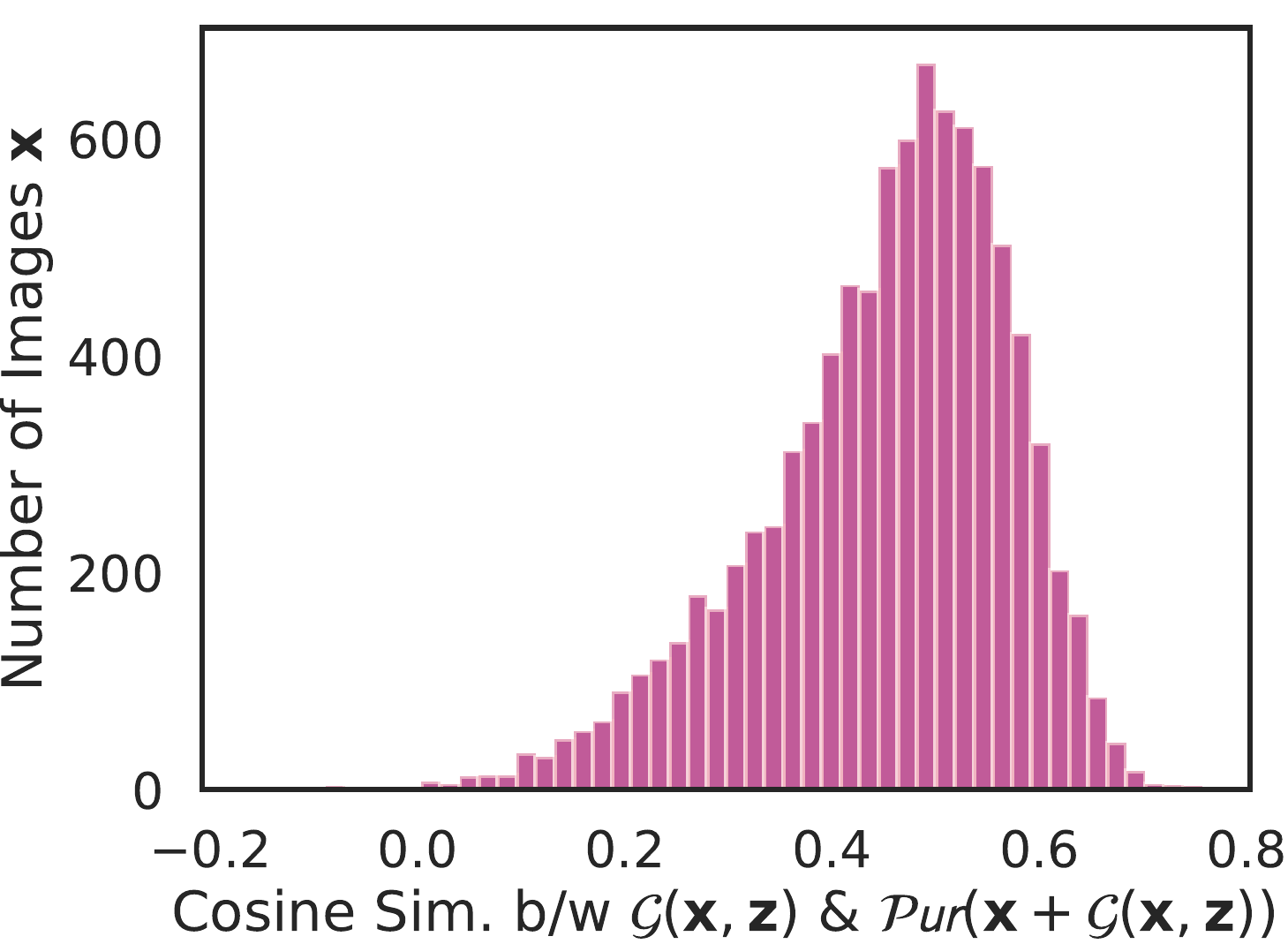}\label{fig:cos_sim}}\hfill
\subfloat[]{\includegraphics[width=0.5\linewidth]{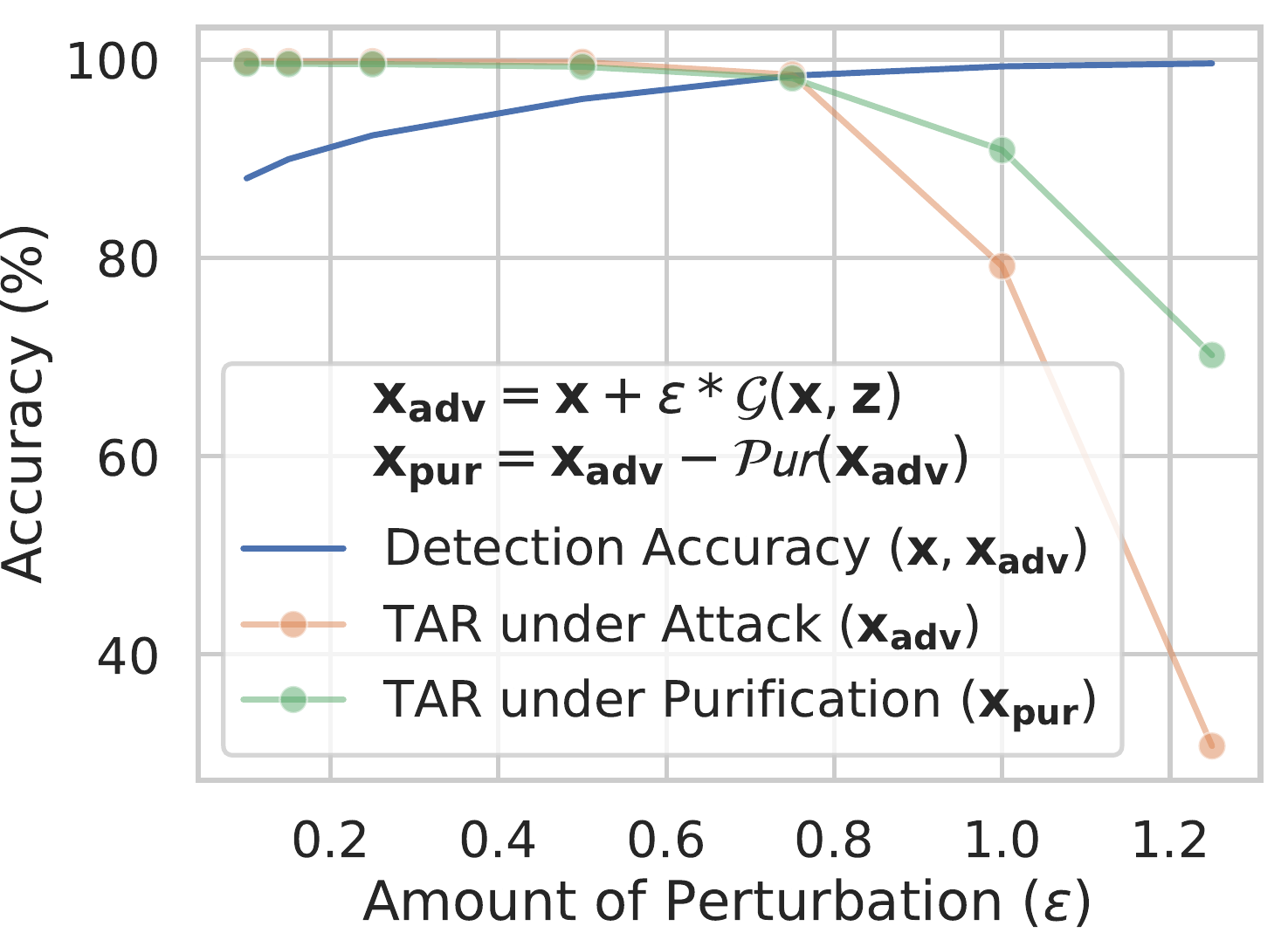}\label{fig:eps}}\vspace{-3mm}
\caption{(a) \emph{FaceGuard}'s purification is correlated with its adversarial synthesis process. (b) Trade-off between detection and purification with respect to perturbation magnitudes. With minimal perturbation, detection is challenging while purifier maintains AFR performance. Excessive perturbations lead to easier detection with greater challenge in purification.}
\label{fig:perturbation_vary}
\end{figure}

\Paragraph{Quality of the Adversarial Purifier} Recall that we enforced the purified image to be close to the real face via a reconstruction loss. Thus, the purification and perturbation masks should be similar. In Fig.~\ref{fig:cos_sim}, we shows that the two masks are indeed correlated by plotting the Cosine similarity distribution ($\in[-1,1]$) between $\EG(\bx, \bz)$ and $\EPur(\bx+\EG(\bx, \bz))$ for all $9,164$ images in LFW. 

Therefore, pixels in $\bx_{adv}$ involved in the purification process should correspond to those that cause the image to be adversarial in the first place. Fig.~\ref{fig:qual_results} highlights that perturbed regions can be automatically localized via constructing a heatmap out of $\EPur(\bx_{adv})$. In Fig.~\ref{fig:eps}, we investigate the change in AFR performance (TAR (\%) @ $0.1$\% FAR) of ArcFace under attack (synthesized adversarial faces via $\EG(\bx,\bz)$) when the amount of perturbation is varied. We find that (a) minimal perturbation is harder to detect but the purifier incurs minimal damage to the AFR, while, (b) excessive perturbations are easier to detect but increases the challenge in purification.

\Section{Conclusions}
With the introduction of sophisticated adversarial attacks on AFR systems, such as geometric warping and GAN-synthesized adversarial attacks, adversarial defense needs to be robust and generalizable. Without utilizing any pre-computed training samples from known adversarial attacks, the proposed~\emph{FaceGuard} achieved state-of-the-art detection performance against $6$ different adversarial attacks. \emph{FaceGuard}'s purifier also enhanced ArcFace's recognition performance under adversarial attacks. We are exploring whether an attention mask predicted by the detector can further improve adversarial purification.

\appendix

\section{Implementation Details}
All the models in the paper are implemented using Tensorflow r1.12. A single NVIDIA GeForce GTX 2080Ti GPU is used for training~\emph{FaceGuard} on CASIA-Webface~\cite{casia} and evaluated on LFW~\cite{lfw}, CelebA~\cite{celeba}, and FFHQ~\cite{ffhq}.
\textbf{Code, pre-trained models and dataset will be publicly available.}

\subsection{Preprocessing}
All face images are first passed through MTCNN
face detector~\cite{mtcnn} to detect $5$ facial landmarks (two eyes,
nose and two mouth corners). Then, similarity transformation is used to normalize the face images based on the five
landmarks. After transformation, the images are resized to
$160\times 160$. Before passing into~\emph{FaceGuard}, each pixel in the
RGB image is normalized $\in [-1, 1]$ by subtracting $128$ and dividing by $128$. \textbf{All the testing images in the main paper and
this supplementary material are from the identities in
the test dataset.}

\subsection{Network Architectures}
The generator, $\EG$ takes as input an real RGB face image, $\bx \in \IR^{160\times160\times3}$ and a $128$-dimensional random latent vector, $\bz \sim \mathcal{N}(0, \mathbf{I})$ and outputs a synthesized adversarial face $\bx_{adv} \in \IR^{160\times160\times3}$. 
Let \texttt{c7s1-k} be a $7\times7$ convolutional layer with $k$ filters and stride $1$. \texttt{dk} denotes a $4\times 4$ convolutional layer with $k$ filters and stride $2$. \texttt{Rk} denotes a residual block that contains two $3\times 3$ convolutional layers. \texttt{uk} denotes a $2\times$ upsampling layer followed by a $5\times 5$ convolutional layer with $k$ filters and stride $1$. We apply Instance Normalization and Batch Normalization to the generator and discriminator, respectively. We use Leaky ReLU with slope $0.2$ in the discriminator and ReLU activation in the generator. The architectures of the two modules are as follows:
\begin{itemize}
    \itemsep0em
    \item Generator: \\ 
    \texttt{c7s1-64,d128,d256,R256,R256,R256, u128, u64, c7s1-3},
    \item Discriminator: \\ \texttt{d32,d64,d128,d256,d512}.
\end{itemize}
A $1\times 1$ convolutional layer with $3$ filters and stride $1$ is attached to the last convolutional layer of the discriminator for the patch-based GAN loss $\EL_{GAN}$.

The purifier, $\EPur$, consists of the same network architecture as the generator:
\begin{itemize}
    \itemsep0em
    \item Purifier: \\ 
    \texttt{c7s1-64,d128,d256,R256,R256,R256, u128, u64, c7s1-3}.
\end{itemize}

\noindent We apply the \texttt{tanh} activation function on the last convolution layer of the generator and the purifier to ensure that the generated images are $\in[-1, 1]$. In the paper, we denoted the output of the tanh layer of the generator as an ``perturbation mask'', $\mathcal{G}(\bx, \bz) \in [-1,1]$ and $\bx \in [-1,1]$. Similarly, the output of the tanh layer of the purifier is referred to an ``purification mask'', $\EPur(\bx_{adv}) \in [-1,1]$ and $\bx_{adv} \in [-1,1]$. The final adversarial image is computed as $\bx_{adv} = 2 \times \texttt{clamp}\left[ \mathcal{G}(\bx,\bz) + \left(\frac{\bx+1}{2}\right) \right]_{0}^{1} -1$. This ensures $\mathcal{G}(\bx, \bz)$ can either add or subtract pixels from $x$ when $\mathcal{G}(\bx, \bz) \neq 0$. When $\mathcal{G}(\bx, \bz)\to0$, then $\bx_{adv}\to \bx$. Similarly, the final purified image is computed as $x_{pur} = 2 \times \texttt{clamp}\left[\left(\frac{\bx_{adv}+1}{2}\right) - \EPur(\bx_{adv})\right]_{0}^{1} -1$.

The external critic network, detector $\ED$, comprises of a $4$-layer binary CNN:
\begin{itemize}
    \itemsep0em
    \item Detector: \\ 
    \texttt{d32,d64,d128,d256,fc64,fc1},
\end{itemize}
where $\texttt{fcN}$ refers to a fully-connected layer with $N$ neuron outputs.

\subsection{Training Details}
The generator, detector, and purifier are trained in an end-to-end manner via ADAM optimizer with hyperparameters $\beta_1 = 0.5$, $\beta_2 = 0.9$, learning rate of $1e-4$, and batch size $16$. Algorithm~\ref{alg} outlines the training algorithm.

\begin{algorithm}[!t]
    \caption{Training \emph{FaceGuard}. All experiments in this work use $\alpha = 0.0001$, $\beta_1 = 0.5$, $\beta_2 = 0.9$, $\lambda_{obf} = \lambda_{fr} = 10.0$, $\lambda_{pt} = \lambda_{perc} = \lambda_{div} = 1.0$, $\epsilon=3.0$, $m = 16$. For brevity, $lg$ refers to log operation.}\label{alg:advfaces}
  \begin{algorithmic}[1]
     \Input
     \Desc{$\mathcal{X}$}{Training Dataset}
      \Desc{$\mathcal{F}$}{Cosine similarity by AFR}
      \Desc{$\mathcal{G}$}{Generator with weights $\mathcal{G}_\theta$}
      \Desc{$Dc$}{Discriminator with weights $Dc_\theta$}
      \Desc{$\mathcal{D}$}{Detector with weights $\mathcal{D}_\theta$}
      \Desc{$\EPur$}{Purifier with weights $\EPur_\theta$}
      \Desc{$m$}{Batch size}
      \Desc{$\alpha$}{Learning rate}
  \EndInput
    \For{number of training iterations}
        \State \text{Sample a batch of probes $\{x^{(i)}\}_{i=1}^{m} \sim \mathcal{X}$}
        \State Sample a batch of random latents $\{z^{(i)}\}_{i=1}^{m} \sim \mathcal{N}(0,I)$
        \State $\delta_\EG^{(i)} = \mathcal{G}((x^{(i)}, z^{(i)})$
        \State $x_{adv}^{(i)} = x^{(i)} + \delta_\EG^{(i)}$
        \State $\delta_{\EPur}^{(i)} = \mathcal{G}((x^{(i)}, z^{(i)})$
        \State $x_{pur}^{(i)} = x_{adv}^{(i)} - \delta_{\EPur}^{(i)}$\\
        \State $\EL^{\mathcal{G}}_{pt} = \frac{1}{m}\left[ \sum_{i=1}^{m}\max\left(\epsilon,||\delta^{(i)}||_2\right)\right]$
        \State $\EL^{\mathcal{G}}_{obf} = \frac{1}{m}\left[\sum_{i=1}^{m}\mathcal{F}\left(x^{(i)}, x_{adv}^{(i)}\right)\right]$
        \State $\EL^{\mathcal{G}}_{div} = -\frac{1}{m}\left[\sum_{i=1}^{m}\left[\frac{\left|\left|\EG(\bx, \bz_1)^{(i)} - \EG(\bx, \bz_2)^{(i)}\right|\right|_1}{\left|\left|\bz_1 - \bz_2\right|\right|_1}\right]\right]$
        \State $\EL^{\mathcal{G}}_{GAN} = \frac{1}{m}\left[\sum_{i=1}^{m}lg\left(1-Dc(x_{adv}^{(i)}) \right)\right]$
        \State $\EL_{\ED} =  \frac{1}{m}\sum_{i=1}^{m}\left[lg\ED(\bx^{(i)}) + lg\left(1 - \ED(\bx_{adv}^{(i)})\right)\right]$
        \State $\EL_{Dc} = \frac{1}{m}\sum_{i=1}^{m}\left[lg\left(Dc(x^{(i)}) \right) +  lg\left(1-Dc(x_{adv}^{(i)}) \right)\right]$
        \State $\EL^{\EPur}_{perc} = \frac{1}{m} \sum_{i=1}^{m}\left[||x_{pur} - x||_1 + ||\EPur(x_{adv}^{(i)})||_1\right]$
        \State $\EL^{\EPur}_{fr} = -\frac{1}{m}\left[\sum_{i=1}^{m}\mathcal{F}\left(x^{(i)}, x_{pur}\right)\right]$
        \State $\EL^{\EPur}_{bf} = \frac{1}{m}\left[\sum_{i=1}^{m}lg\left(1-\mathcal{D}(x_{pur}) \right)\right]$
        \State $\EL_{\mathcal{G}} = \EL^{\mathcal{G}}_{GAN} + \lambda_{obf} \EL_{obf} + \lambda_{pt} \EL_{pt} + \lambda_{div}\EL_{div}$
        \State $\EL_{\EPur} = \lambda_{fr} \EL_{fr} + \lambda_{perc} \EL_{perc} + \lambda_{bf}\EL_{bf}$
        \State $\mathcal{G}_{\theta} = \texttt{Adam}(\triangledown_{\mathcal{G}}\EL^{\mathcal{G}}, \mathcal{G}_{\theta},\alpha,\beta_{1},\beta_{2})$
        \State $Dc\theta = \texttt{Adam}(\triangledown_{Dc}\EL^{Dc}, Dc_{\theta},\alpha,\beta_{1},\beta_{2})$
        \State $\mathcal{D}_\theta = \texttt{Adam}(\triangledown_{\mathcal{D}}\EL^{\mathcal{D}}, \mathcal{D}_{\theta},\alpha,\beta_{1},\beta_{2})$
        \State $\EPur_\theta = \texttt{Adam}(\triangledown_{\EPur}\EL^{\EPur}, \EPur_{\theta},\alpha,\beta_{1},\beta_{2})$
      \EndFor
    \end{algorithmic}
    \label{alg}
\end{algorithm}

\begin{figure}[!t]
    \centering
    \captionsetup{font=footnotesize}
    \includegraphics[width=0.8\linewidth]{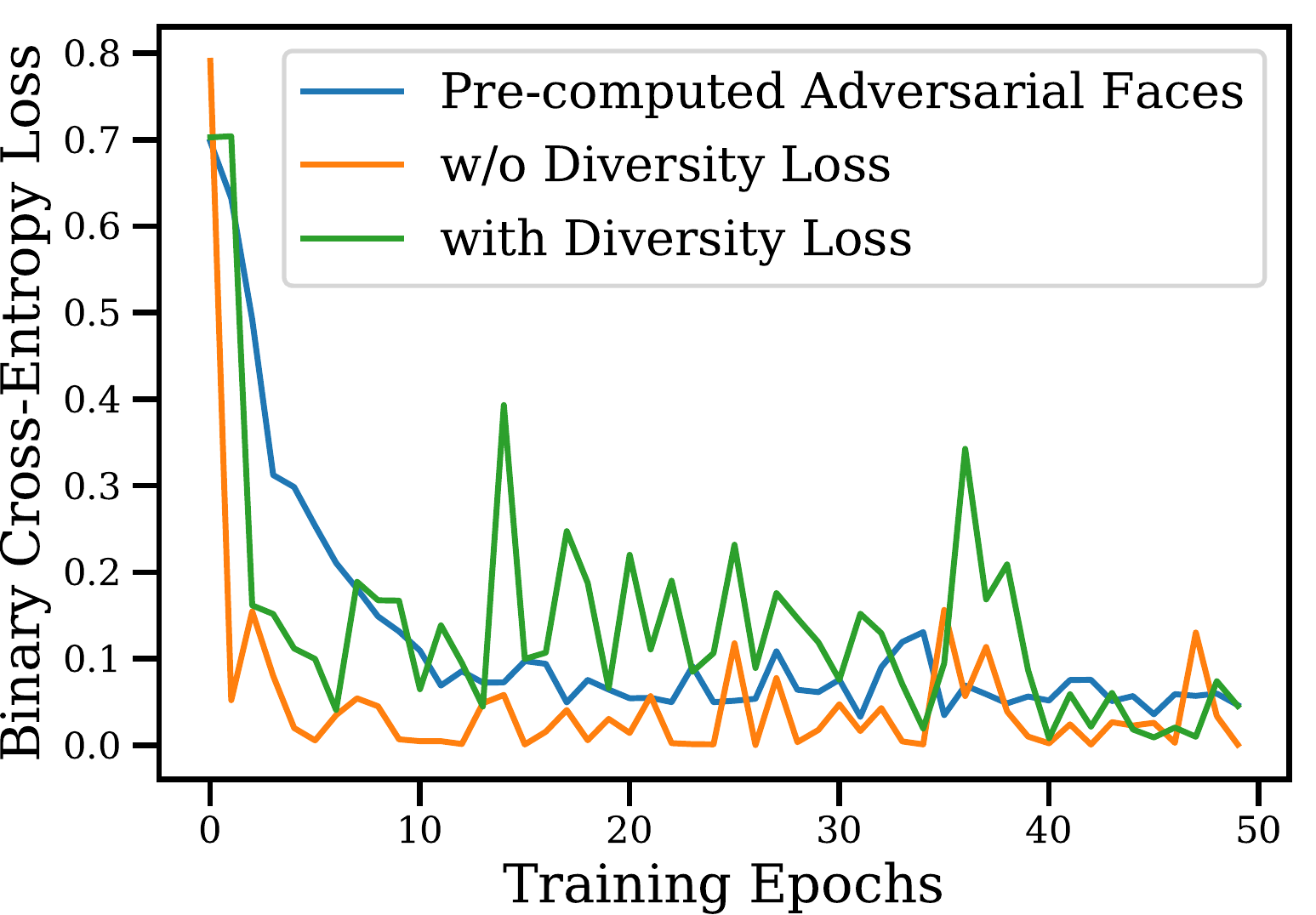}
    \caption{Training loss across iterations when an adversarial detection network is trained via pre-computed adversarial faces (blue), the proposed adv. generator but without the diversity (orange), and with the proposed diversity loss (green). The diversity loss prevents the network from overfitting to adversarial perturbations encountered during training.}
    \label{fig:loss}
\end{figure}

\Paragraph{Network Convergence}  In Fig.~\ref{fig:loss}, we plot the training loss across iterations when an adversarial detector is trained via pre-computed adversarial faces. In this case, the training loss converges to a low value and remains consistent across the remaining epochs. Such a detector may overfit to the fixed set of adversarial perturbations encountered in training (see Supp.). Instead of utilizing the pre-computed adversarial attacks, utilizing an adversarial generator in training (without $\EL_{div}$), introduces challenging training samples.

\emph{FaceGuard} with the diversity loss introduces diverse perturbations within a training iteration (see Fig.~\ref{fig:diverse}; main paper). In Fig.~\ref{fig:loss}, we also observe that the training loss significantly fluctuates (epochs $8-40$) until convergence (epochs $40-50$). This indicates that throughout the training (within and across training iterations), the proposed generator synthesizes strong and diverse range of adversarial faces that continuously regularizes the training of the adversarial detector.

\begin{table*}[!t]
\centering
\ra{1.0}
\footnotesize
\captionsetup{font=footnotesize}
\begin{threeparttable}
\begin{tabular}{l|lMMMM}
\toprule \multicolumn{1}{l}{} & \textbf{Detection Accuracy (\%)} & \textbf{Year} & \textbf{LFW~\cite{fgsm}} & \textbf{CelebA~\cite{celeba}} & \textbf{FFHQ~\cite{ffhq}}\\
\midrule
\parbox[t]{2mm}{\multirow{7}{*}{\raisebox{5.5em}{\rotatebox[origin=c]{90}{General}}}}
& Gong~\etal\cite{gong} & 2017 & 97.54 \pm 02.82 & 94.38 \pm 04.48 & 96.89 \pm 02.07\\
& ODIN~\cite{odin} & 2018 & 77.03 \pm 14.34 & 68.95 \pm 19.64 & 74.63 \pm 08.16\\
& Steganalysis~\cite{steganalysis} & 2019 & 74.33 \pm 14.77 & 72.53 \pm 11.30 & 71.09 \pm 09.86\\
\midrule
\parbox[t]{2mm}{\multirow{11}{*}{\raisebox{5.5em}{\rotatebox[origin=c]{90}{Face}}}}
& UAP-D~\cite{uapd} & 2018 & 64.28 \pm 09.97 & 63.19 \pm 16.49 & 68.65 \pm 08.73\\
& SmartBox~\cite{smartbox} & 2018 & 56.77 \pm 05.16 & 54.85 \pm 09.33 & 57.19 \pm 09.55\\
& Goswami~\etal\cite{goswami2019detecting} & 2019 & 79.37 \pm 14.04 & 74.70 \pm 13.88 & 80.03 \pm 09.24\\
& Massoli~\etal\cite{massoli}~(MLP) & 2020 & 69.16 \pm 15.29 & 61.78 \pm 11.34 & 66.26 \pm 10.06\\
& Massoli~\etal\cite{massoli}~(LSTM) & 2020 & 70.11 \pm 13.35 & 63.67 \pm 16.21 & 69.58 \pm 07.91\\
& Agarwal~\etal\cite{agarwal_image_transform} & 2020 & 87.03 \pm 16.86 & 85.81 \pm 15.64 & 86.70 \pm 11.04\\
\midrule
\multicolumn{1}{l}{} & \emph{Proposed FaceGuard} & 2021 & \mathbf{99.81 \pm 00.10} & \mathbf{98.73 \pm 00.92} & \mathbf{99.35 \pm 00.09}\\
\bottomrule
\end{tabular}
\end{threeparttable}\vspace{-2mm}
\caption{Average and standard deviation of detection accuracies of SOTA adversarial face detectors in classifying six adversarial attacks synthesized for the LFW~\cite{lfw}, CelebA~\cite{celeba}, and FFHQ~\cite{ffhq} datasets. Detection threshold is set as $0.5$ for all methods. All baseline methods require training on pre-computed adversarial attacks on CASIA-WebFace~\cite{casia}. On the other hand, the proposed~\emph{FaceGuard} is self-guided and generates adversarial attacks on the fly. Hence, it can be regarded as a~\emph{black-box} defense system.}
\label{tab:detection_celeba}
\end{table*}

\begin{table*}[!t]
\centering
\footnotesize
\captionsetup{font=footnotesize}
\begin{threeparttable}
\begin{tabular}{l|MMM|MMM}
\noalign{\hrule height 1.5pt}
 & \multicolumn{3}{c|}{\textbf{Known}} & \multicolumn{3}{c}{\textbf{Unseen}}\\
\noalign{\hrule height 0.5pt}
& \textbf{FGSM~\cite{fgsm}} & \textbf{PGD~\cite{pgd}} & \textbf{DeepFool~\cite{deepfool}} & \textbf{AdvFaces~\cite{advfaces}} & \textbf{GFLM~\cite{gflm}} & \textbf{SemanticAdv~\cite{semantic_adv}}\\
\noalign{\hrule height 1.0pt}
Gong~\etal\cite{gong} & 94.51 & 92.21 & 94.12 & 68.63 & 50.00 & 50.21\\
UAP-D~\cite{uapd} & 63.65 & 69.33 & 56.38 & 60.81 & 50.12 & 50.28\\
SmartBox~\cite{smartbox} & 58.79 & 62.53 & 51.32 & 54.87 & 50.97 & 62.14\\
Massoli~\etal\cite{massoli}~(MLP) & 78.35 & 82.52 & 91.21 & 55.57 & 50.00 & 50.00\\
Massoli~\etal\cite{massoli}~(LSTM) & 74.61 & 86.43 & 94.73 & 62.43 & 50.00 & 50.00\\
\noalign{\hrule height 1.0pt}
\end{tabular}
\end{threeparttable}\\\vspace{1mm}{\centering (a)}\\\vspace{2mm}
\begin{threeparttable}
\begin{tabular}{l|MMM|MMM}
\noalign{\hrule height 1.5pt}
 & \multicolumn{3}{c|}{\textbf{Known}} & \multicolumn{3}{c}{\textbf{Unseen}}\\
\noalign{\hrule height 0.5pt}
& \textbf{AdvFaces~\cite{advfaces}} & \textbf{GFLM~\cite{gflm}} & \textbf{SemanticAdv~\cite{semantic_adv}} & \textbf{FGSM~\cite{fgsm}} & \textbf{PGD~\cite{pgd}} & \textbf{DeepFool~\cite{deepfool}}\\
\noalign{\hrule height 1.0pt}
Gong~\etal\cite{gong} & 81.39 & 96.72 & 98.97 & 84.46 & 57.00 & 72.32\\
UAP-D~\cite{uapd} & 68.78 & 54.31 & 77.46 & 51.64 & 50.32 & 52.01\\
SmartBox~\cite{smartbox} & 54.87 & 50.97 & 62.14 & 58.79 & 62.53 & 51.32\\
Massoli~\etal\cite{massoli}~(MLP) & 77.64 & 86.54 & 94.78 & 55.20 & 51.32 & 52.90\\
Massoli~\etal\cite{massoli}~(LSTM) & 81.42 & 92.62 & 96.76 & 52.74 & 65.43 & 54.84\\
\noalign{\hrule height 1.0pt}
\end{tabular}
\end{threeparttable}\\\vspace{1mm}{\centering (b)}\\\vspace{2mm}
\begin{threeparttable}
\begin{tabular}{l|MMMMMM}
\noalign{\hrule height 1.5pt}
 & \multicolumn{6}{c}{\textbf{Known}}\\\noalign{\hrule height 1.0pt}
 & \textbf{FGSM~\cite{fgsm}} & \textbf{PGD~\cite{pgd}} & \textbf{DeepFool~\cite{deepfool}} & \textbf{AdvFaces~\cite{advfaces}} & \textbf{GFLM~\cite{gflm}} & \textbf{SemanticAdv~\cite{semantic_adv}}\\
\noalign{\hrule height 0.5pt}
Gong~\etal\cite{gong} & 98.94 & 97.91 & 95.87 & 92.69 & \mathbf{99.92} & \mathbf{99.92}\\
UAP-D~\cite{uapd} & 61.32 & 74.33 & 56.78 & 51.11 & 65.33 & 76.78\\
SmartBox~\cite{smartbox} & 58.79 & 62.53 & 51.32 & 54.87 & 50.97 & 62.14\\
Massoli~\etal\cite{massoli}~(MLP) & 63.58 & 76.28 & 81.78 & 88.38 & 51.97 & 52.98\\
Massoli~\etal\cite{massoli}~(LSTM) & 71.53 & 76.43 & 88.32 & 75.43 & 53.76 & 55.22\\
\noalign{\hrule height 0.5pt}
& \multicolumn{6}{c}{\textbf{Unseen}}\\
\noalign{\hrule height 0.5pt}
\emph{Proposed FaceGuard} & \mathbf{99.85} & \mathbf{99.85} & \mathbf{99.85} & \mathbf{99.84} & 99.61 & 99.85\\
\noalign{\hrule height 1.0pt}
\end{tabular}
\end{threeparttable}\\\vspace{1mm}{\centering (c)}
\caption{Detection accuracy of SOTA adversarial face detectors in classifying six adversarial attacks synthesized for the LFW dataset~\cite{lfw} under various known and unseen attack scenarios. Detection threshold is set as $0.5$ for all methods.}
\label{tab:detection_scenarios}
\end{table*}

\SubSection{Baselines} We evaluate all defense methods via publicly available repositories provided by the authors. Only modification made is to replace their training datasets with CASIA-WebFace~\cite{casia}. We provide the public links to the author codes below:
\begin{itemize}[noitemsep]
    \item Gong~\etal\cite{gong}: \url{https://github.com/gongzhitaao/adversarial-classifier}
    \item UAP-D\cite{uapd}/SmartBox~\etal\cite{smartbox}: \url{https://github.com/akhil15126/SmartBox}
    \item Massoli~\etal\cite{massoli}: \url{https://github.com/fvmassoli/trj-based-adversarials-detection}
    \item Adversarial Training~\cite{adv_train}:
    \url{https://github.com/locuslab/fast_adversarial}
    \item Rob-GAN~\cite{robgan}:
    \url{https://github.com/xuanqing94/RobGAN}
    \item L2L~\cite{l2l}: \url{https://github.com/YunseokJANG/l2l-da}
    \item MagNet~\cite{magnet}:
    \url{https://github.com/Trevillie/MagNet}
    \item DefenseGAN~\cite{defense_gan}:
    \url{https://github.com/kabkabm/defensegan}
    \item NRP~\cite{self_supervised}:
    \url{https://github.com/Muzammal-Naseer/NRP}
\end{itemize}
Attacks are also synthesized via publicly available author codes:
\begin{itemize}[noitemsep]
    \item FGSM/PGD/DeepFool: \url{https://github.com/tensorflow/cleverhans}
    \item AdvFaces: \url{https://github.com/ronny3050/AdvFaces}
    \item GFLM: \url{https://github.com/alldbi/FLM}
    \item SemanticAdv: \url{https://github.com/AI-secure/SemanticAdv}
\end{itemize}

\section{Additional Datasets}
In Tab.~\ref{tab:detection_celeba}, we report average and standard deviation of detection rates of the proposed~\emph{FaceGuard} and other baselines on the $6$ adversarial attacks synthesized on LFW~\cite{lfw}, CelebA~\cite{celeba}, and FFHQ~\cite{ffhq} (following the same protocol as Tab.~\ref{tab:detection} in main paper). For CelebA, we synthesize a total of $19,962\times6 = 119,772$ adversarial samples for $19,962$ real samples in the CelebA testing split~\cite{celeba}. We also synthesize $4,974\times6=29,844$ adversarial samples for $4,974$ real faces in FFHQ testing split~\cite{ffhq}.  We find that the proposed~\emph{FaceGuard} outperforms all baselines in all three face datasets.

\section{Overfitting in Prevailing Detectors}
In Tab.~\ref{tab:detection_scenarios}, we provide the detection rates of prevailing SOTA detectors in detecting six adversarial attacks in LFW~\cite{lfw} when they are trained on different attack subsets. We highglight the overfitting issue when (a) SOTA detectors are trained on gradient-based adversarial attacks (FGSM~\cite{fgsm}, PGD~\cite{pgd}, and DeepFool~\cite{deepfool}) and tested on gradient-based and learning-based attacks (AdvFaces~\cite{advfaces}, GFLM~\cite{gflm}, and SemanticAdv~\cite{semantic_adv}), and (b) vice-versa. Tab.~\ref{tab:detection_scenarios}(c) reports the detection performance of SOTA detectors when all six attacks are available for training.

We find that detection accuracy of SOTA detectors significantly drops when tested on a subset of attacks not encountered during their training. Instead, the proposed~\emph{FaceGuard} maintains robust detection accuracy without even training on the pre-computed samples from any known attacks.

\begin{figure*}[!t]
    \centering
    \captionsetup{font=footnotesize}
    \includegraphics[width=\linewidth]{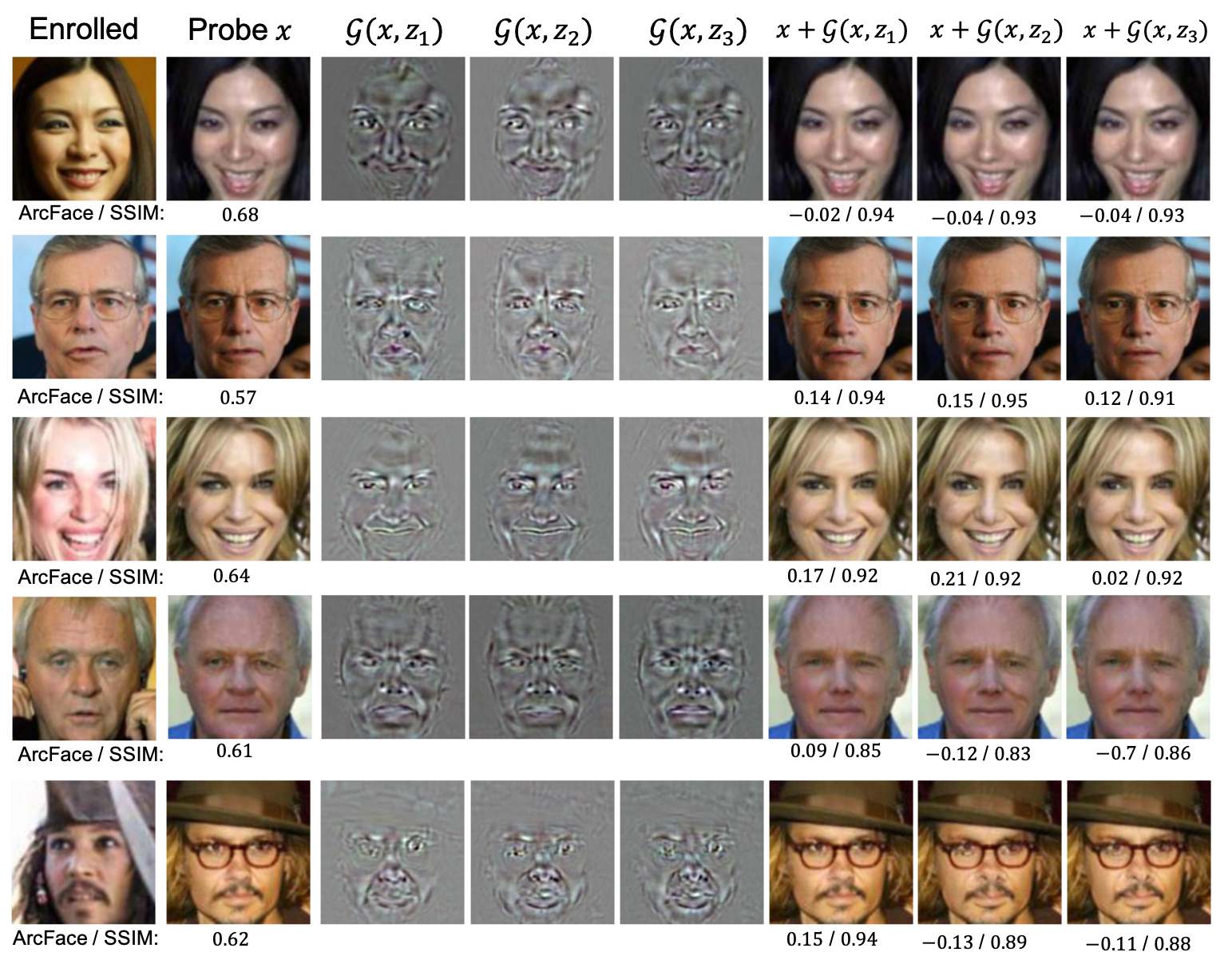}
    \caption{Examples of generated adversarial images along with corresponding perturbation masks obtained via~\emph{FaceGuard}'s generator $\EG$ for three randomly sampled $\bz$. Cosine similarity scores via ArcFace~\cite{arcface} $\in [-1,1]$ and SSIM $\in [0,1]$ between synthesized adversarial and input probe are given below each image. A score above $\textbf{0.36}$ (threshold @ $0.1\%$ False Accept Rate) indicates that two faces are of the same subject.}
    \label{fig:gen}
\end{figure*}

\Section{Qualitative Results}
\SubSection{Generator Results}
Fig.~\ref{fig:gen} shows examples of synthesized adversarial faces via the proposed adversarial generator $\EG$. Note that the generator takes the input prob $\bx$ and a random latent $\bz$. We show synthesized perturbation masks and corresponding adversarial faces for three randomly sampled latents. We observe that the synthesized adversarial images evades ArcFace~\cite{arcface} while maintaining high structural similarity between adversarial and input probe. 

\begin{figure*}[!t]
    \centering
    \captionsetup{font=footnotesize}
    \includegraphics[width=0.9\linewidth]{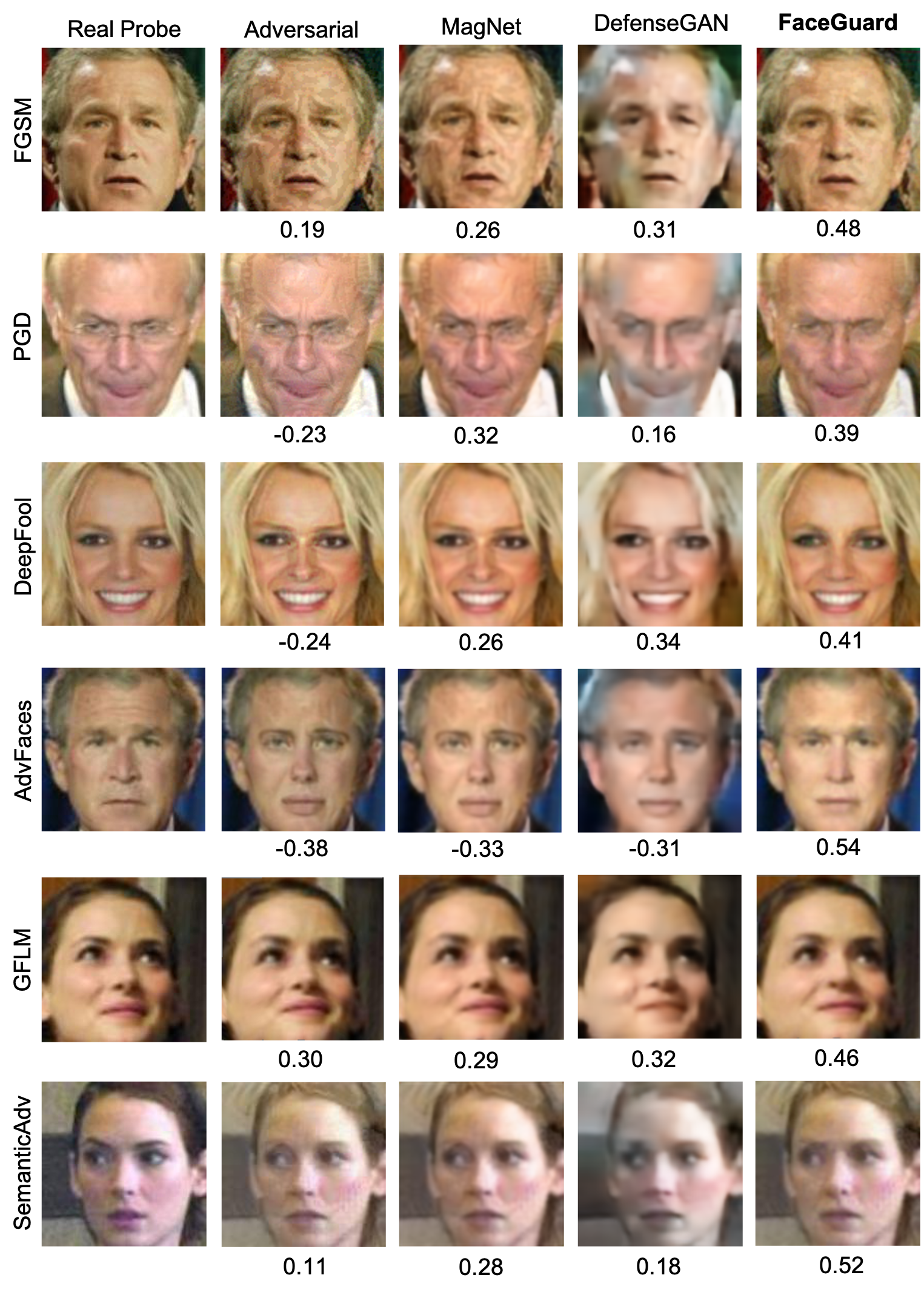}\vspace{-4mm}
    \caption{Examples of purified images via MagNet~\cite{magnet}, DefenseGan~\cite{defense_gan}, and proposed~\emph{FaceGuard} purifiers for six adversarial attacks. Cosine similarity scores via ArcFace~\cite{arcface} $\in [-1,1]$ are given below each image. A score above $\textbf{0.36}$ (threshold @ $0.1\%$ False Accept Rate) indicates that two faces are of the same subject.}
    \label{fig:pur}
\end{figure*}

\SubSection{Purifier Results}
We show examples of purified images via~\emph{FaceGuard} and baselines including MagNet~\cite{magnet} and DefenseGAN~\cite{defense_gan} in Fig.~\ref{fig:pur}. We observe that, compared to baselines, purified images synthesized via~\emph{FaceGuard} are visually realistic with minimal changes compared to the ground truth real probe. In addition, compared to the two baselines, ~\emph{FaceGuard}'s purifier protects ArcFace~\cite{arcface} matcher from being evaded by the six adversarial attacks.

\begin{figure*}[!t]
    \centering
    \footnotesize
    \captionsetup{font=footnotesize}
    \includegraphics[width=\linewidth]{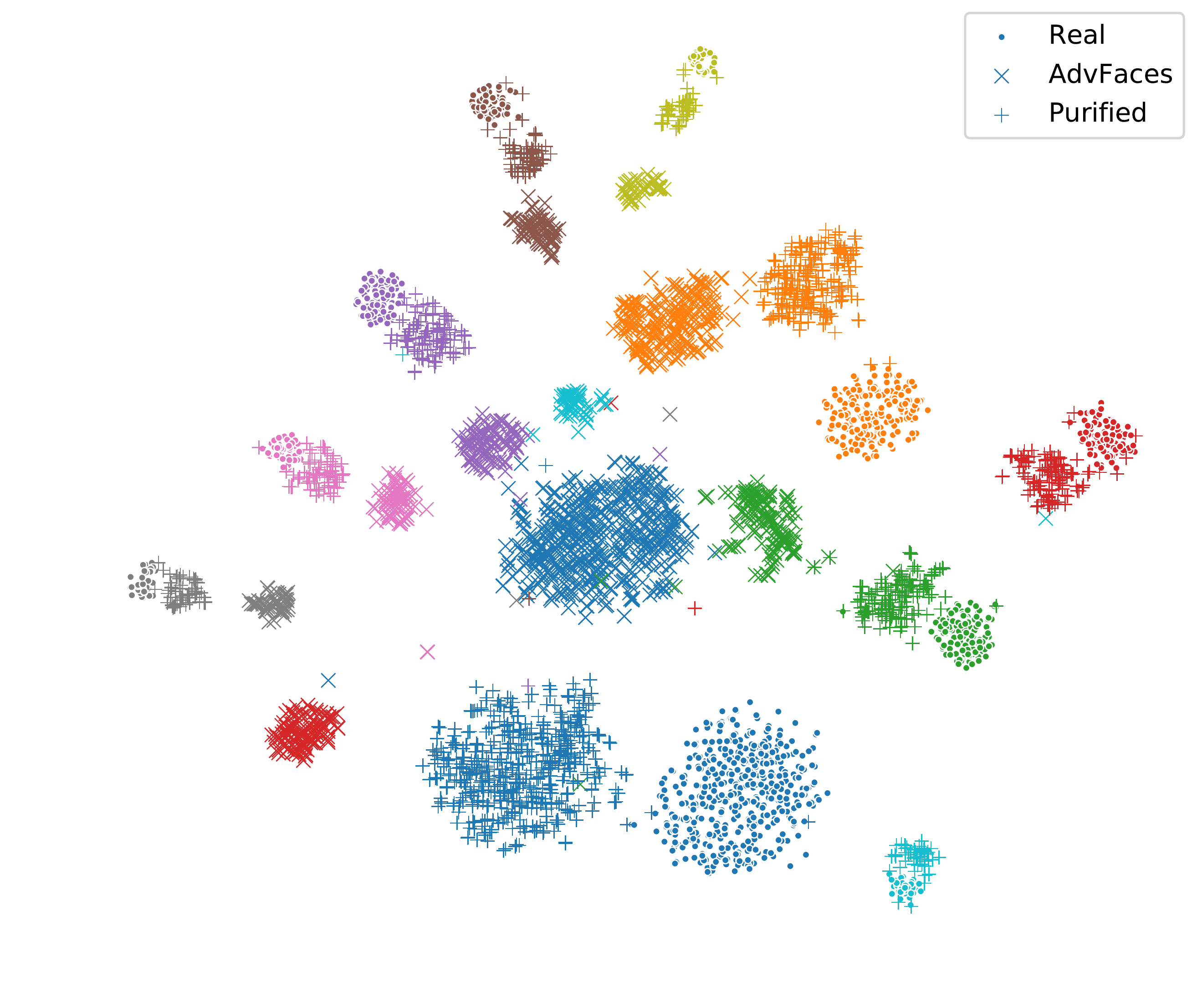}
    \caption{$2$D t-SNE visualization of face representations extracted via ArcFace from $1,456$ (a) real, (b) AdvFaces~\cite{advfaces}, and (c) purified images belonging to $10$ subjects in LFW~\cite{lfw}. Example AdvFaces~\cite{advfaces} pertaining to a subject moves farther from its identity cluster while the proposed purifier draws them back.}
    \label{fig:tsne}
\end{figure*}

\Section{Additional Results on Purifier}
\SubSection{Perturbation and Purification Masks}
In the main text, we found that the perturbation and purification masks are correlated with an average Cosine similarity of $0.52$. We show five pairs of perturbation and purification masks ranked by the Cosine similarity between them (highest to lowest). We observe that purification mask is better correlated when perturbations are more local. Slightly perturbing entire faces poses to be challenging for the proposed purifier.
\SubSection{Effect of Perturbation Amount} We also studied the effect of perturbation amount on detection and purification results in the main text. We observed a trade-off between detection and purification with
respect to perturbation magnitudes. With minimal perturbation, detection
is challenging while purifier maintains AFR performance. Excessive perturbations lead to easier detection with greater challenge in purification. In Fig.~\ref{fig:eps}, show examples of synthesized adversarial faces for different perturbation amounts and their corresponding purified images. We find that detection scores improve with larger perturbation. Aligned with our earlier findings, due to the proposed bonafide loss, $\EL_{bf}$, purified faces are continuously detected as real by the detector which explains why the purifier maintains AFR performance with increasing perturbation amount.
\SubSection{Effect of Purification on ArcFace Embeddings}
In order to investigate the effect of purification on a matcher's feature space, we extract face embeddings of real images, their corresponding adversarial images via the challenging AdvFaces~\cite{advfaces} attack, and purified images, via the SOTA ArcFace matcher. In total, we extract feature vectors from $1,456$ face images of $10$ subjects in the LFW dataset~\cite{lfw}. In Fig.~\ref{fig:tsne}, we plot the $2$D t-SNE visualization of the face embeddings for the $10$ subjects. The identity clusterings can be clearly observed from real, adversarial, and purified images. In particular, we observe that some adversarial faces pertaining to a subject moves farther from its identity cluster while the proposed purifier draws them back. Fig.~\ref{fig:tsne} illustrates that the proposed purifier indeed enhances face recognition performance of ArcFace under attack from $34.27\%$ TAR @ $0.1\%$ FAR under no defense to $77.46\%$ TAR @ $0.1\%$ FAR.
\begin{figure*}[!t]
    \centering
    \footnotesize
\captionsetup{font=footnotesize}
    \includegraphics[width=\linewidth]{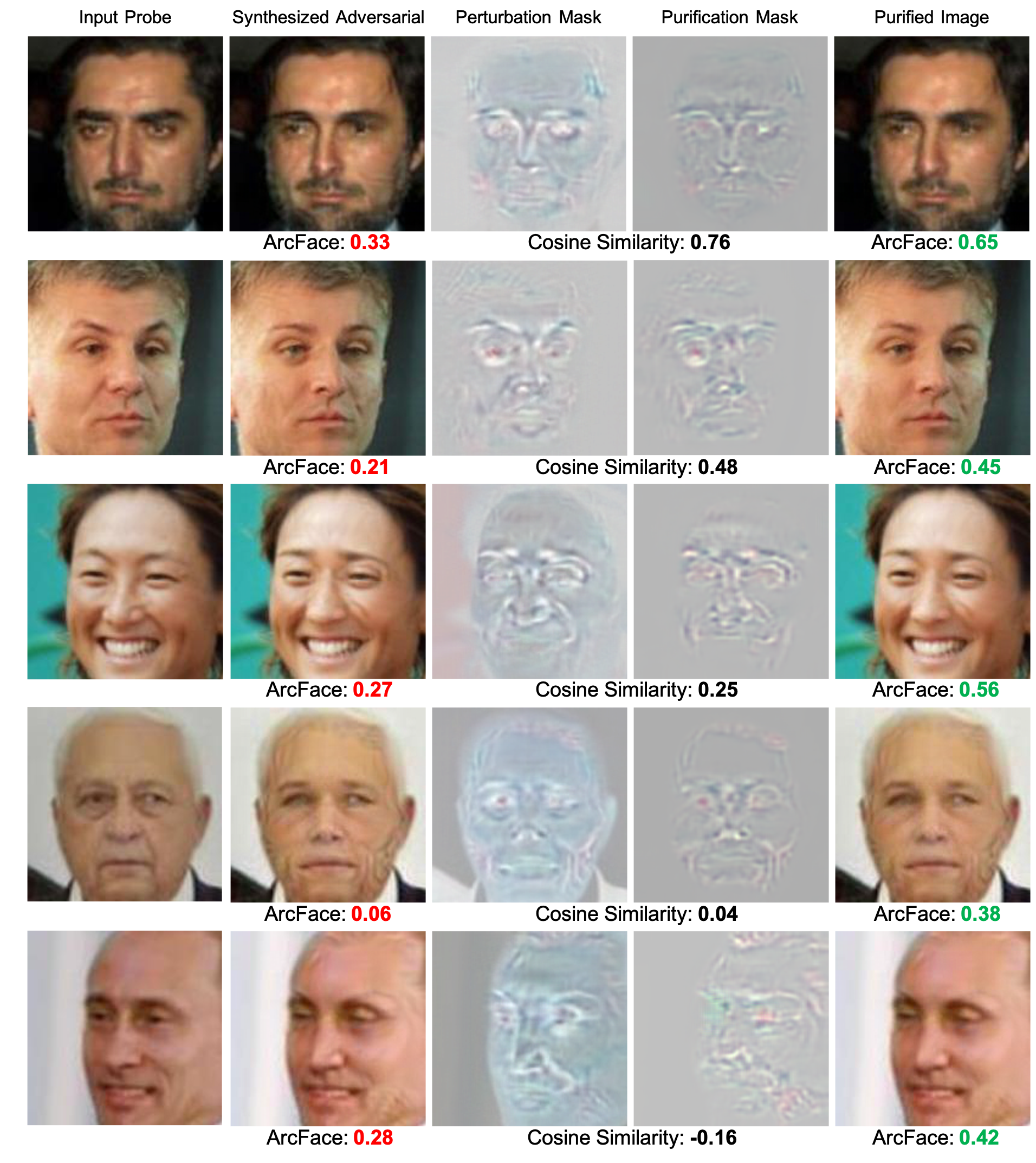}\vspace{-4mm}
    \caption{Examples of synthesized adversarial images via the proposed adversarial generator and corresponding purified images. Cosine similarity between perturbation and purification masks given below each row along with ArcFace scores between synthesized adversarial/purified image and real probe.  A score above $\textbf{0.36}$ (threshold @ $0.1\%$ False Accept Rate) indicates that two faces are of the same subject. Even with lower correlation between perturbation and purification masks (rows 3-5), the purified images can still be identified as the correct identity. Notice that the purifier primarily alters the eye color, nose, and subdues adversarial perturbations in foreheads. Zoom in for details.}
    \label{fig:gxpx}
\end{figure*}

\begin{figure*}[!t]
    \centering
    \footnotesize
\captionsetup{font=footnotesize}
    \includegraphics[width=\linewidth]{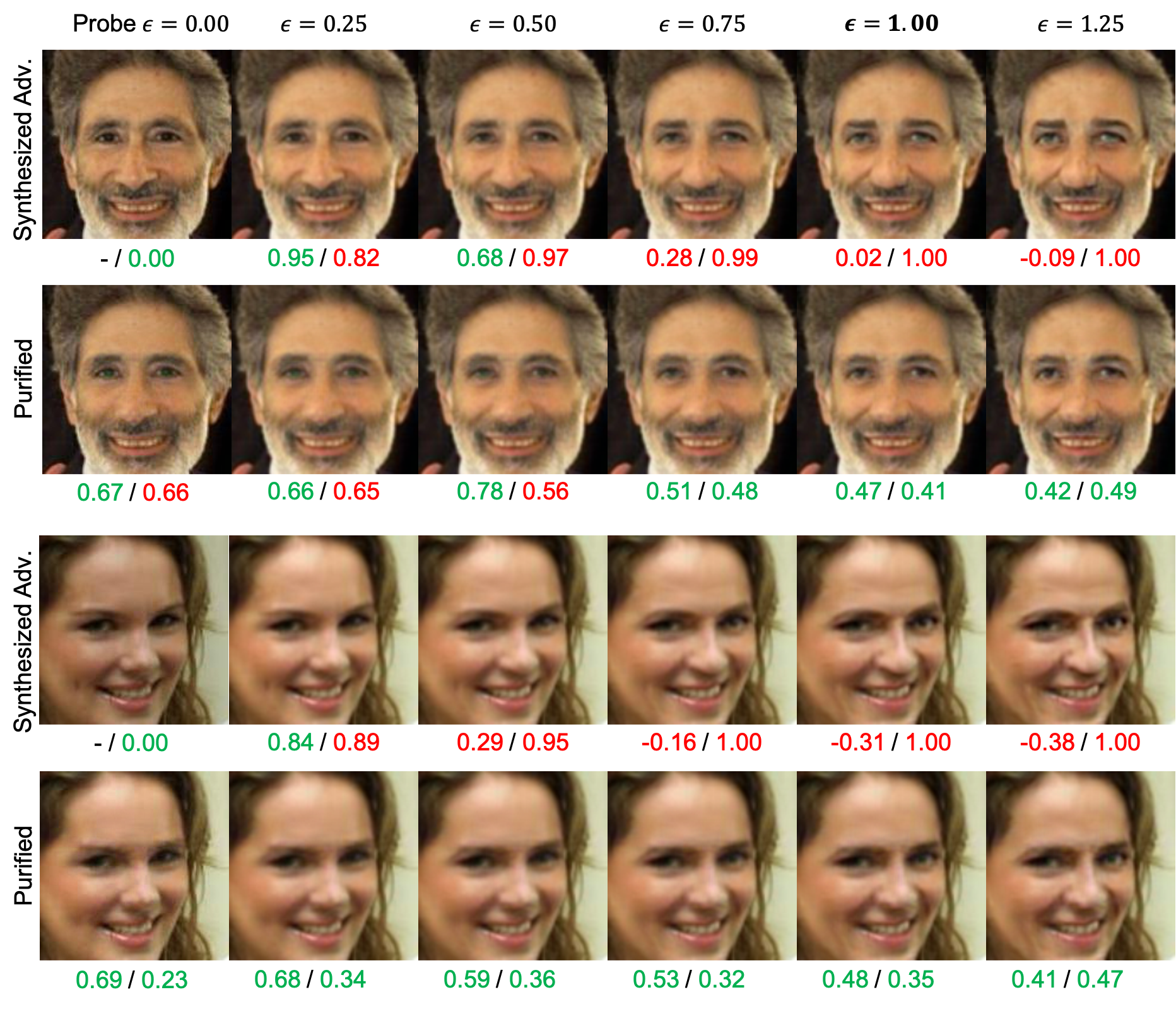}\vspace{-4mm}
    \caption{ArcFace $\in [-1, 1]$ / Detection scores $\in [0, 1]$ when perturbation amount is varied ($\epsilon = \{0.25, 0.50, 0.75, 1.00, 1.25\}$). Detection scores above $0.5$ are predicted as adversarial images while ArcFace scores above $\textbf{0.36}$ (threshold @ $0.1\%$ False Accept Rate) indicate that two faces are of the same subject. \emph{FaceGuard} is trained on $\epsilon=1.00$. The detection scores improve as perturbation amount increases, whereas, majority of purified images are detected as real. Even when purified images fail to be classified as real by the detector, purification maintain high AFR performance.}
    \label{fig:eps}
\end{figure*}

\clearpage
{
\small
\bibliographystyle{unsrt}
\bibliography{egbib}
}

\end{document}